\pdfoutput=1

\documentclass[11pt]{article}

\usepackage{emnlp2021}

\usepackage{times}
\usepackage{latexsym}
\usepackage{microtype}
\usepackage{booktabs}
\usepackage{multirow}
\usepackage{amsmath}
\usepackage{amssymb}
\usepackage{array}
\usepackage{xspace}
\usepackage{latexsym}
\usepackage{graphicx}
\usepackage{dblfloatfix}
\usepackage[linesnumbered,ruled,vlined]{algorithm2e}
\usepackage[T1]{fontenc}

\usepackage[utf8]{inputenc}

\usepackage{graphicx}
\usepackage[inline]{enumitem}
\usepackage{amsmath}
\usepackage{amssymb}
\usepackage{booktabs}
\usepackage{bm}
\usepackage{array}
\usepackage{xcolor}
\usepackage{stmaryrd}
\usepackage{scalerel}
\usepackage{multirow}
\usepackage{capt-of,tabularx}
\newcolumntype{C}{>{\centering\arraybackslash}X}

\newcommand{\ourtag}[1]{\ensuremath{\left\langle #1 \right\rangle}}

\newcommand{\shufmodel}[3]{\ensuremath{\text{#1}^{\text{#3}}}}

\makeatletter
\newcommand*{\bigcdot}{}
\DeclareRobustCommand*{\bigcdot}{%
  \mathbin{\mathpalette\bigcdot@{}}%
}
\newcommand*{\bigcdot@scalefactor}{.65}
\newcommand*{\bigcdot@widthfactor}{1.15}
\newcommand*{\bigcdot@}[2]{%
  \sbox0{$#1\vcenter{}$}
  \sbox2{$#1\cdot\m@th$}%
  \hbox to \bigcdot@widthfactor\wd2{%
    \hfil
    \raise\ht0\hbox{%
      \scalebox{\bigcdot@scalefactor}{%
        \lower\ht0\hbox{$#1\bullet\m@th$}%
      }%
    }%
    \hfil
  }%
}
\makeatother

\newcommand{\sep}{$\bigcdot$}
\newcommand{\vertmulticell}[2]{\multirow{#1}{*}{\rotatebox[origin=c]{90}{#2}}}

\usepackage{microtype}

\title{Investigating Pretrained Language Models for Graph-to-Text Generation}

\author{Leonardo F. R. Ribeiro$^{\dag}$, Martin Schmitt$^{\ddag}$, Hinrich Sch{\"u}tze$^{\ddag}$ and Iryna Gurevych$^{\dag}$ \vspace{1mm} \\
\rule{0pt}{2.5ex}
  $^{\dag}$Research Training Group AIPHES and UKP Lab, Technical University of Darmstadt\\
  $^{\ddag}$Center for Information and Language Processing (CIS), LMU Munich \\
  \rule{0pt}{2.5ex}
 \texttt{\href{https://www.ukp.tu-darmstadt.de}{www.ukp.tu-darmstadt.de}}
}

\begin{document}
\maketitle
\begin{abstract}
Graph-to-text generation aims to generate fluent texts from graph-based data. In this paper, we investigate two recent pretrained language models (PLMs) and analyze the impact of different task-adaptive pretraining strategies for PLMs in graph-to-text generation. We present a study across three graph domains: meaning representations, Wikipedia knowledge graphs (KGs) and scientific KGs. We show that approaches based on PLMs BART and T5 achieve new state-of-the-art results and that task-adaptive pretraining strategies improve their performance even further. We report new state-of-the-art BLEU scores of 49.72 on AMR-LDC2017T10, 59.70 on WebNLG, and 25.66 on AGENDA datasets - a relative improvement of 31.8\%, 4.5\%, and 42.4\%, respectively, with our models generating significantly more fluent texts than human references.  In an extensive analysis, we identify possible reasons for the PLMs’ success on graph-to-text tasks. Our findings suggest that the PLMs benefit from similar facts seen during pretraining or fine-tuning, such that they perform well even when the input graph is reduced to a simple bag of node and edge labels.\footnote{Our code is available at \href{https://github.com/UKPLab/plms-graph2text}{https://github.com/UKPLab/plms-graph2text}.}
\end{abstract}

\section{Introduction}

Graphs are important data structures in NLP as they represent complex relations within a set of objects. For example, semantic and syntactic structures of sentences can be represented using different graph representations (e.g., AMRs, \citeauthor{banarescu-etal-2013-abstract}, \citeyear{banarescu-etal-2013-abstract}; semantic-role labeling, \citeauthor{surdeanu-etal-2008-conll}, \citeyear{surdeanu-etal-2008-conll}; syntactic and semantic graphs, \citeauthor{belz-etal-2011-first}, \citeyear{belz-etal-2011-first}) and knowledge graphs (KGs) are used to describe factual knowledge in the form of relations between entities \cite{gardent-etal-2017-webnlg, VOUGIOUKLIS20181, koncel-kedziorski-etal-2019-text}.

\begin{figure*}[t]
    \centering
    \includegraphics[width=0.9\textwidth]{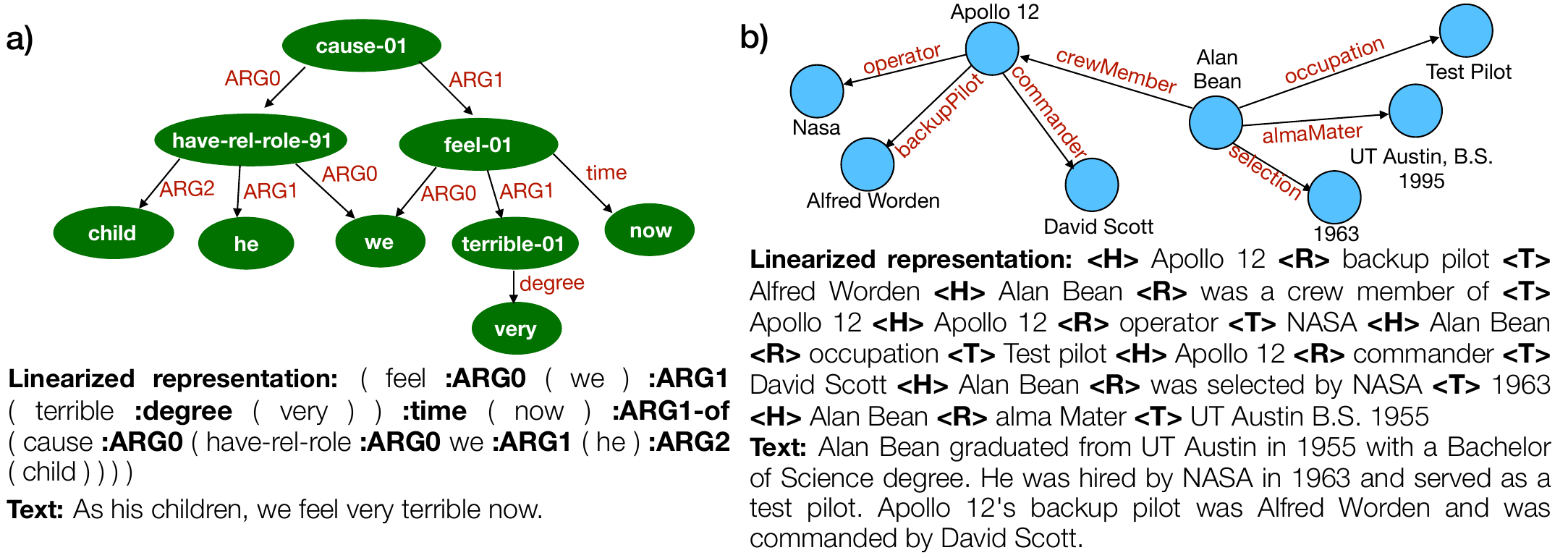}
    \caption{Examples of (a) AMR and (b) WebNLG graphs, the input for the models and the reference texts.}
    \label{fig:graphs}
\end{figure*}

Graph-to-text generation, a subtask of data-to-text generation \cite{10.5555/3241691.3241693}, aims to create fluent natural language text to describe an input graph (see Figure~\ref{fig:graphs}). This task is important for NLP applications such as dialogue generation \cite{moon-etal-2019-opendialkg} and question answering \cite{duan-etal-2017-question}. Recently, it has been shown that structured meaning representation, such as AMR or KG, can store the internal state of a dialog system, providing core semantic knowledge \cite{bonial-etal-2020-dialogue, bai-etal-2021-semantic} or can be the result of a database query for conversational QA \cite{yu-etal-2019-cosql}. Moreover, dialog states can be represented as KGs to encode compositionality and can be shared across different domains, slot types and dialog participators \cite{cheng-etal-2020-conversational}.

Transfer learning has become ubiquitous in NLP and pretrained Transformer-based architectures \cite{NIPS2017_7181} have considerably outperformed prior state of the art in various downstream tasks \cite{devlin-etal-2019-bert, NEURIPS2019_dc6a7e65, liu2020roberta, radford2019language}. 


In this paper, we analyze the applicability of two recent text-to-text pretrained language models (PLMs), BART \cite{lewis2019bart} and T5 \cite{2019t5}, for graph-to-text generation. We choose these models because of their \emph{encoder-decoder} architecture, which makes them particularly suitable for conditional text generation. Our study comprises three graph domains (meaning representations, Wikipedia KGs, and scientific KGs). We further introduce \emph{task-adaptive} graph-to-text pretraining approaches for PLMs and demonstrate that such strategies improve the state of the art by a substantial margin.

While recent works have shown the benefit of explicitly encoding the graph structure in graph-to-text generation \cite[][to name a few]{song-etal-acl2018, ribeiro-etal-2019-enhancing, ribeiro-etal-2020-modeling,schmitt2020modeling, zhao-etal-2020-bridging}, our approaches based on PLMs consistently outperform these models, even though PLMs -- as sequence models -- do not exhibit any \emph{graph-specific structural bias}.\footnote{The model architecture does not explicitly encode the graph structure, i.e., which entities are connected to each other, but has to retrieve it from a sequence that tries to encode this information.} Simply representing the graph as a linear traversal (see Figure~\ref{fig:graphs}) leads to remarkable generation performance in the presence of a strong language model.
In our analysis
we investigate to what extent fine-tuned PLMs make use of the graph structure represented in the graph linearization.
We notably observe that PLMs achieve high performance on two popular KG-to-text benchmarks even when the KG is reduced to a mere bag of node and edge labels.

Our contributions are the following:
\begin{itemize}[noitemsep,nolistsep]
    \item We investigate and compare two PLMs, BART and T5, for graph-to-text generation, exploring \emph{language model adaptation} (\textsc{lma}) and \emph{supervised task adaptation} (\textsc{sta}) pretraining, employing additional task-specific data.
    \item Our approaches consistently outperform the state of the art by significant margins, ranging from 2.6 to 12.0 BLEU points, on three established graph-to-text benchmarks from different domains, exceeding specialized graph architectures (e.g., Graph Neural Networks, GNNs, \citeauthor{Kipf:2016tc}, \citeyear{Kipf:2016tc}).
    \item In a crowdsourcing experiment, we demonstrate that our methods generate texts with significantly better fluency than existing works and the human references.
    \item We discover that PLMs perform well even when trained on a shuffled linearized graph representation without any information about connectivity (bag of node and edge labels), which is surprising since prior studies showed that explicitly encoding the graph structure improves models trained from scratch (e.g., \citeauthor{zhao-etal-2020-bridging}, \citeyear{zhao-etal-2020-bridging}); and investigate the possible reasons for such a good performance.
    
    
\end{itemize}

\section{Related Work}

\paragraph{Graph-to-text Learning.} Various neural models have been proposed to generate sentences from graphs from different domains. \citet{konsas_17} propose the first neural approach for AMR-to-text generation that uses a linearized input graph. Prior approaches for KG-to-text generation train text-to-text neural models using sequences of KG triples as input \cite{trisedya-etal-2018-gtr, moryossef-etal-2019-step, castro-ferreira-etal-2019-neural, ribeiro2021smelting}. 

Recent approaches \cite{marcheggiani-icnl18, song-etal-acl2018, beck-etal-2018-acl2018, damonte-cohen-2019-structural, ribeiro-etal-2019-enhancing,zhao-etal-2020-bridging, schmitt-etal-2021-modeling, ribeiro2021structural} propose architectures based on GNNs to directly encode the graph structure, whereas other efforts \cite{ribeiro-etal-2020-modeling, schmitt2020modeling, yao-etal-2020-heterogeneous, doi:10.116200297} inject the graph structure information into Transformer-based architectures. The success of those approaches suggests that imposing a strong relational inductive bias into the graph-to-text model can assist the generation.

\paragraph{Pretrained Language Models.}  Pretrained Transformer-based models, such as BERT \cite{devlin-etal-2019-bert}, XLNet \cite{NIPS2019_8812}, or RoBERTa \cite{liu2020roberta}, have established a qualitatively new level of baseline performance for many widely used natural language understanding (NLU) benchmarks. Generative pretrained Transformer-based methods, such as GPT-2 \cite{radford2019language}, BART \cite{lewis2019bart}, and T5 \cite{2019t5}, are employed in many natural language generation (NLG) tasks.  

\citet{mager2020gpttoo} were the first to employ GPT-2, a decoder-only PLM, for AMR-to-text generation and use cycle consistency to improve the adequacy. In contrast, we are the first to investigate BART and T5 models, which have both a Transformer-based encoder and decoder, in AMR-to-text generation. Recently, \citet{harkous2020text} and \citet{kale2020texttotext} demonstrate state-of-the-art results in different data-to-text datasets, employing GPT-2 and T5 models respectively. \citet{radev2020dart} propose DART, a new data-to-text dataset, and train a BART model gradually augmenting the WebNLG training data with DART data. 

\citet{hoyle2020promoting} explore scaffolding objectives in PLMs and show gains in low-resource graph-to-text settings.
Different from the above works, we focus on a general transfer learning strategies for graph-to-text generation, investigating task-adaptive pretraining approaches, employing additional collected task-specific data for different PLMs (BART and T5) and benchmarks. In addition, we provide a detailed analysis aimed at explaining the good performance of PLMs on KG-to-text tasks.

Recently, \citet{gururangan-etal-2020-dont} explored task-adaptive pretraining strategies for text classification.
While our \textsc{lma} (see \S\ref{sec:finetuning}) is related to their \textsc{dapt} as both use a self-supervised objective on a domain-specific corpus,
they notably differ in that \textsc{dapt} operates on the model input while \textsc{lma} models the output. We are the first to show the benefits of additional task-specific pretraining in PLMs for graph-to-text tasks.


\section{PLMs for Graph-to-Text Generation}
\label{sec:finetuning}
\subsection{Models in this Study}
We investigate BART \cite{lewis2019bart} and T5 \cite{2019t5}, two PLMs based on the Transformer \emph{encoder-decoder} architecture \cite{NIPS2017_7181}, for graph-to-text generation. 
They mainly differ in how they are pretrained and the input corpora used for pretraining. We experiment with different T5 (\emph{small} - 60M  parameters, \emph{base} - 220M, and \emph{large} - 770M) and BART (\emph{base} - 140M and \emph{large} - 400M) capacity models.

We fine-tune both PLMs for a few epochs on the supervised downstream graph-to-text datasets. For T5, in the supervised setup, we add a prefix ``translate from Graph to Text:'' before the graph input. We add this prefix to imitate the T5 setup, when translating between different languages. 



\subsection{Task-specific Adaptation} 
\label{sec:domainpretraining}


Inspired by previous work \cite{konsas_17,gururangan-etal-2020-dont}, we investigate whether leveraging additional task-specific data can improve the PLMs' performance on graph-to-text generation. Task-specific data refers to a pretraining corpus that is more task-relevant and usually smaller than the text corpora used for task-independent pretraining. In order to leverage the task-specific data, we add an intermediate adaptive pretraining step between the original pretraining and fine-tuning phases for graph-to-text generation. 


More precisely, we first continue pretraining BART and T5 using language model adaptation (\textsc{lma}) or supervised task adaptation (\textsc{sta}) training. In the supervised approach, we use pairs of graphs and corresponding texts collected from the same or similar domain as the target task. In the \textsc{lma} approach, we follow BART and T5 pretraining strategies for language modeling, using the reference texts that describe the graphs. Note that we do not use the graphs in the \textsc{lma} pretraining, but only the target text of our task-specific data collections. The goal is to adapt the decoder to the domain of the final task  \cite{gururangan-etal-2020-dont}. In particular, we randomly mask text spans, replacing 15\% of the tokens.\footnote{Please, refer to \citet{lewis2019bart} and \citet{2019t5} for details about the self-supervised pretraining strategies.}
Before evaluation, we finally fine-tune the models using the original training set as usual.

\section{Datasets}
\label{sec:data}

We evaluate the text-to-text PLMs on three graph-to-text benchmarks: AMR (LDC2017T10), WebNLG \cite{gardent-etal-2017-webnlg}, and AGENDA \cite{koncel-kedziorski-etal-2019-text}. We chose those datasets because they comprise different domains and are widely used in prior work. Table~\ref{tab:datastatistics} in Appendix shows statistics for each dataset.

\paragraph{AMR.} Abstract meaning representation (AMR) is a semantic formalism that represents the meaning of a sentence as a rooted directed graph expressing ``who is doing what to whom'' \cite{banarescu-etal-2013-abstract}. In an AMR graph, nodes represent concepts and edges represent semantic relations. An instance in LDC2017T10 consists of a sentence annotated with its corresponding AMR graph. Following ~\citet{mager2020gpttoo}, we linearize the AMR graphs using the \textsc{penman} notation (see Figure~\ref{fig:graphs}a).\footnote{Details of the preprocessing procedure of AMRs are provided in Appendix~\ref{sec:amrinput}.}



\paragraph{WebNLG.} Each instance of WebNLG contains a KG from DBPedia \cite{10.5555/1785162.1785216} and a target text with one or multiple sentences that describe the graph. The test set is divided into two partitions: \textit{seen}, which contains only DBPedia categories present in the training set, and \textit{unseen}, which covers categories never seen during training. Their union is called \textit{all}.
Following previous work \cite{harkous2020text}, we prepend \ourtag{H}, \ourtag{R}, and \ourtag{T} tokens before the head entity, the relation and tail entity of a triple (see Figure~\ref{fig:graphs}b).

\paragraph{AGENDA.} In this dataset, KGs are paired with scientific abstracts extracted from proceedings of AI conferences. Each sample contains the paper title, a KG, and the corresponding abstract. The KG contains entities corresponding to scientific terms and the edges represent relations between these entities. This dataset has loose alignments between the graph and the corresponding text as the graphs were automatically generated. The input for the models is a text containing the title, a sequence of all KG entities, and the triples. The target text is the paper abstract. We add special tokens into the triples in the same way as for WebNLG. 

\subsection{Additional Task-specific Data}
In order to evaluate the proposed task-adaptive pretraining strategies for graph-to-text generation, we collect task-specific data for two graph domains: meaning representations (like AMR) and scientific data (like AGENDA).
We did not attempt collecting additional data like WebNLG because the texts in this benchmark do not stem from a corpus but were specifically written by annotators.

\paragraph{AMR Silver Data.} In order to generate additional data for AMR, we sample two sentence collections of size 200K and 2M from the Gigaword\footnote{\href{https://catalog.ldc.upenn.edu/LDC2003T05}{https://catalog.ldc.upenn.edu/LDC2003T05}} corpus and use a state-of-the-art AMR parser \cite{cai-lam-2020-amr} to parse them into AMR graphs.\footnote{We filter out sentences that do not yield well-formed AMR graphs.} For supervised pretraining, we condition a model on the AMR silver graphs to generate the corresponding sentences before fine-tuning it on gold AMR graphs. For self-supervised pretraining, we only use the sentences.\footnote{Gigaword and AMR datasets share similar data sources.}

\begin{table}[t]
\centering
{\renewcommand{\arraystretch}{0.6}
\begin{tabular}{@{\hspace*{1mm}}l@{\hspace*{1mm}}ccc@{\hspace*{1mm}}}  
\toprule
\textbf{Model} & \textbf{BLEU} & \textbf{M} & \textbf{BT}  \\
\midrule
\citet{ribeiro-etal-2019-enhancing}  & 27.87 & 33.21 & -\\
\citet{zhu-etal-2019-modeling} & 31.82 & 36.38 & -\\
\citet{zhao-etal-2020-line} & 32.46 & 36.78 & -\\
\citet{doi:10.116200297} & 33.90 & 37.10 &  -\\
\citet{yao-etal-2020-heterogeneous}  & 34.10 & 38.10 & -\\
\midrule
\small{\textit{based on PLMs}} & & & \\[.2em]
\citet{mager2020gpttoo}  & 33.02 & 37.68 &  -\\
\citet{harkous2020text}  & 37.70 & 38.90 &-\\
\midrule
BART\textsubscript{base}  & 36.71 & 38.64 & 52.47 \\
BART\textsubscript{large}  & 43.47 & 42.88 &
60.42 \\
T5\textsubscript{small} & 38.45 & 40.86 & 57.95 \\
T5\textsubscript{base} & 42.54 & 42.62 & 60.59 \\
T5\textsubscript{large} & \textbf{45.80} & \textbf{43.85} & \textbf{61.93} \\
\midrule
\multicolumn{3}{l}{\small{\textit{with task-adaptive pretraining}}} & \\[.2em]
BART\textsubscript{large} + \textsc{lma} & 43.94 & 42.36 & 58.54 \\
T5\textsubscript{large} + \textsc{lma} & 46.06 & 44.05 & 62.59 \\[.7em]

BART\textsubscript{large} + \textsc{sta} \small{\textsc{(200K)}} & 44.72 & 43.65 & 61.03 \\
BART\textsubscript{large} + \textsc{sta} \small{\textsc{(2M)}} & 47.51 & 44.70 & 62.27 \\
T5\textsubscript{large} + \textsc{sta} \small{\textsc{(200K)}} & 48.02 & 44.85 & 63.86 \\
T5\textsubscript{large} + \textsc{sta} \small{\textsc{(2M)}} & \textbf{\textit{49.72}} & \textbf{\textit{45.43}} & \textbf{\textit{64.24}} \\
\bottomrule
\end{tabular}}
\vspace{-2mm}
\caption{Results on AMR-to-text generation for the LDC2017T10 test set. M and BT stand for METEOR and BLEURT, respectively. \textbf{Bold} (\textbf{\textit{Italic}}) indicates the best score without (with) task-adaptive pretraining.}
\label{tab:results-amr}
\vspace{-0.2cm}
\end{table}

\begin{table*}[t]
\small
\centering
{\renewcommand{\arraystretch}{0.9}
\begin{tabular}{@{\hspace*{1mm}}l@{\hspace*{3mm}}ccccccccc@{\hspace*{1mm}}}  
\toprule
& \multicolumn{3}{c}{\textbf{BLEU}} & \multicolumn{3}{c}{\textbf{METEOR}} & \multicolumn{3}{c}{\textbf{chrF++}} \\
\midrule
\textbf{Model} & \textbf{A} & \textbf{S} & \textbf{U} & \textbf{A} & \textbf{S} & \textbf{U} & \textbf{A} & \textbf{S} & \textbf{U} \\
\midrule
\citet{castro-ferreira-etal-2019-neural} & 51.68 & 56.35 & 38.92 & 32.00 & 41.00 &21.00  & - & - & - \\
\citet{moryossef-etal-2019-step} & 47.24 & 53.30 & 34.41 & 39.00 & 44.00 & 37.00 & - & - & - \\
\citet{schmitt2020modeling} & - & 59.39 & - & - & 42.83 & - & - & 74.68 & - \\
\citet{ribeiro-etal-2020-modeling} & - & 63.69 & - & - & 44.47 & - & - & 76.66 & - \\
\citet{zhao-etal-2020-bridging} & 52.78 & 64.42 & 38.23 & 41.00 & 46.00 & 37.00 & - & - & - \\
\midrule
\multicolumn{2}{l}{\small{\textit{based on PLMs}}} & & \\[.2em]
\citet{harkous2020text} & 52.90 & - & - & 42.40 & - & - & - & - & - \\
\citet{kale2020texttotext} & 57.10 & 63.90 & 52.80 & 44.00 & 46.00 & 41.00 & - & - & -\\
\citet{radev2020dart} & 45.89 & 52.86 & 37.85 & 40.00 & 42.00 & 37.00 & - & - & -\\
\midrule
BART\textsubscript{base} & 53.11 & 62.74 & 41.53 & 40.18 & 44.45 & 35.36 & 70.02 & 76.68 & 62.76 \\
BART\textsubscript{large} & 54.72 & 63.45 & 43.97 & 42.23 & 45.49 & 38.61 & 72.29 & 77.57 & 66.53 \\
T5\textsubscript{small} & 56.34 & \textbf{65.05} & 45.37 & 42.78 & 45.94 & 39.29 & 73.31 & \textbf{78.46} & 67.69 \\
T5\textsubscript{base} & 59.17 & 64.64 & 52.55 & 43.19 & \textbf{46.02} & 41.49 & 74.82 & 78.40 & 70.92 \\
T5\textsubscript{large} & \textbf{59.70} & 64.71 & \textbf{53.67} & \textbf{44.18} & 45.85 & \textbf{42.26} & \textbf{75.40} & 78.29 & \textbf{72.25} \\

\bottomrule
\end{tabular}}
\caption{Results on WebNLG. A, S and U stand for \textit{all}, \textit{seen}, and \textit{unseen} partitions of the test set, respectively.}
\label{tab:results-webnlg}   
\end{table*}

\paragraph{Semantic Scholar AI Data.} We collect titles and abstracts of around 190K scientific papers from the Semantic Scholar \cite{ammar-etal-2018-construction} taken from the proceedings of 36 top Computer Science/AI conferences. We construct KGs from the paper abstracts employing DyGIE++ \cite{wadden-etal-2019-entity}, an information extraction system for scientific texts. Note that the AGENDA dataset was constructed using the older SciIE system \cite{luan-etal-2018-multi}, which also extracts KGs from AI scientific papers. A second difference is that in our new dataset, the domain is broader as we collected data from 36 conferences compared to 12 from AGENDA. Furthermore, to prevent data leakage, all AGENDA samples used for performance evaluation are removed from our dataset. We will call the new dataset \textsc{KGAIA} (KGs from AI Abstracts).\footnote{We will release the collected additional task-specific data.} Table~\ref{tab:augstatistics} in Appendix shows relevant dataset statistics.

\section{Experiments}


We modify the BART and T5 implementations released by Hugging Face \citep{wolf2019huggingfaces} in order to adapt them to graph-to-text generation. For the KG datasets, we add the \ourtag{H}, \ourtag{R}, and \ourtag{T} tokens to the models' vocabulary. We add all edge labels seen in the training set to the vocabulary of the models for AMR. Following \citet{wolf2019huggingfaces}, we use the Adam optimizer \cite{kingma:adam} with an initial learning rate of $3 \cdot 10^{-5}$. We employ a linearly decreasing learning rate schedule without warm-up. The batch and beam search sizes are chosen from \{2,4,8\} and \{1,3,5\}, respectively, based on the respective development set. Dev BLEU is used for model selection.

Following previous works, we evaluate the results with BLEU \cite{Papineni:2002:BMA:1073083.1073135}, METEOR \cite{Denkowski14meteoruniversal}, and chrF++ \cite{popovic-2015-chrf} metrics. We also use MoverScore~\cite{zhao-etal-2019-moverscore}, BERTScore~\cite{bert-score}, and BLEURT~\cite{sellam-etal-2020-bleurt} metrics, as they employ contextual and semantic knowledge and thus depend less on the surface symbols. Additionally, we perform a human evaluation (cf.\ \S\ref{sec:human_eval}) quantifying the fluency, semantic adequacy and meaning similarity of the generated texts.



\subsection{Results on AMR-to-Text}
\label{sec:amr}
Table~\ref{tab:results-amr} shows our results for the setting without additional pretraining, with additional self-supervised task-adaptive pretraining solely using the collected Gigaword sentences (\textsc{lma}), and with additional supervised task adaptation (\textsc{sta}), before fine-tuning. We also report several recent results on the AMR test set. \citet{mager2020gpttoo} and \citet{harkous2020text} employ GPT-2 in their approaches. Note that GPT-2 only consists of a Transformer-based decoder.

Only considering approaches without task adaptation, BART\textsubscript{large} already achieves a considerable improvement of 5.77 BLEU and 3.98 METEOR scores over the previous state of the art. With a BLEU score of 45.80, T5\textsubscript{large} performs best. The other metrics follow similar trends. See Table~\ref{tab:results-amr-appendix} in Appendix for evaluation with more metrics. The strong performance of both BART and T5 in the AMR dataset suggests that PLMs can infer the AMR structure by a simple linear sequence of the graph, in contrast to GNN-based models that explicitly consider the graph structure using \emph{message-passing} between adjacent nodes \cite{beck-etal-2018-acl2018}.

\paragraph{Task-specific Adaptation.}
\textsc{lma}
already brings some gains with T5 benefitting more than BART in most metrics.
It still helps less than \textsc{sta} even though we only have automatically generated annotations.
This suggests that the performance increases with \textsc{sta} do not only come from additional exposure to task-specific target texts and that the models learn how to handle graphs and the graph-text correspondence even with automatically generated AMRs. After \textsc{sta}, T5 achieves 49.72 BLEU points, the new state of the art for AMR-to-text generation. 
Interestingly, gains from \textsc{sta} with 2M over 200K are larger in BART than in T5, suggesting that large amounts of silver data may not be required for a good performance with T5. 

In general, models pretrained on the \textsc{sta} setup converge faster than without task-specific adaptation. For example, T5\textsubscript{large} without additional pretraining converges after 5 epochs of fine-tuning whereas T5\textsubscript{large} with \textsc{sta} already converges after 2 epochs.

\subsection{Results on WebNLG}


Table~\ref{tab:results-webnlg} shows the results for the WebNLG test set. Neural pipeline models \cite{moryossef-etal-2019-step, castro-ferreira-etal-2019-neural} achieve strong performance in the \emph{unseen} dataset. On the other hand, fully end-to-end models \cite{ribeiro-etal-2020-modeling,schmitt2020modeling} have strong performance on the \emph{seen} dataset and usually perform poorly in \textit{unseen} data. Models that \emph{explicitly encode the graph structure} \cite{ribeiro-etal-2020-modeling, zhao-etal-2020-bridging} achieve the best performance among approaches that do not employ PLMs. Note that T5 is also used in \citet{kale2020texttotext}. Differences in our T5 setup include a modified model vocabulary, the use of beam search, the learning rate schedule and the prefix before the input graph. Our T5 approach achieves 59.70, 65.05 and 54.69 BLEU points on \emph{all}, \emph{seen} and \emph{unseen} sets, the new state of the art. 

We conjecture that the performance gap between \emph{seen} and \emph{unseen} sets stems from the advantage obtained by a model seeing examples of relation-text pairs during fine-tuning. For example, the relation \emph{party} (political party) was never seen during training and the model is required to generate a text that verbalizes the tuple: $\langle$\emph{Abdul Taib Mahmud, party, Parti Bumiputera Sarawak}$\rangle$. Interestingly, BART performs much worse than T5 on this benchmark, especially in the \emph{unseen} partition with 9.7 BLEU points lower compared to T5.

For lack of a suitable data source (cf.\ \S\ref{sec:data}), we did not explore our \textsc{lma} or \textsc{sta} approaches for WebNLG.
However, we additionally discuss cross-domain \textsc{sta} in Appendix~\ref{sec:crossdomain}.



\subsection{Results on AGENDA}

\begin{table}[t]
\centering
{\renewcommand{\arraystretch}{0.9}
\begin{tabular}{@{\hspace*{1mm}}lccc@{\hspace*{1mm}}}  
\toprule
\textbf{Model} & \textbf{BLEU} & \textbf{M} & \textbf{BT}  \\
\midrule
Koncel et al. \citeyear{koncel-kedziorski-etal-2019-text}  & 14.30 & 18.80 & - \\
\citet{An2019RepulsiveBS} & 15.10 & 19.50 & -\\
\citet{schmitt2020modeling} & 17.33 & 21.43  & -\\
\citet{ribeiro-etal-2020-modeling} & 18.01 & 22.23 & -\\
\midrule
BART\textsubscript{base} & 22.01 & 23.54 & -13.02 \\
BART\textsubscript{large} & \textbf{23.65} & \textbf{25.19} & \textbf{-10.93} \\
T5\textsubscript{small} & 20.22 & 21.62 & -24.10 \\
T5\textsubscript{base} & 20.73 & 21.88 & -21.03 \\
T5\textsubscript{large} & 22.15 & 23.73 & -13.96 \\
\midrule
\multicolumn{2}{l}{\small{\textit{with task-adaptive pretraining}}} & & \\[.2em]
BART\textsubscript{large} + \textsc{lma} & 25.30 & 25.54 & -08.79 \\
T5\textsubscript{large} + \textsc{lma} & 22.92  & 24.40 & -10.39 \\[.7em]
BART\textsubscript{large} + \textsc{sta} & \textbf{\textit{25.66}} & \textbf{\textit{25.74}} & \textbf{\textit{-08.97}} \\
T5\textsubscript{large} + \textsc{sta} & 23.69 & 24.92 & -08.94 \\
\bottomrule
\end{tabular}}
\caption{Results on AGENDA test set. \textbf{Bold} (\textbf{\textit{Italic}}) indicates best scores without (with) task-adaptive pretraining.}
\label{tab:results-agenda}
\vspace{-0.4cm}
\end{table}

Table~\ref{tab:results-agenda} lists the results for the AGENDA test set. The models also show strong performance on this dataset. We believe that their capacity to generate fluent text helps when generating paper abstracts, even though they were not pretrained in the scientific domain. BART\textsubscript{large} shows an impressive performance with a BLEU score of 23.65, which is 5.6 points higher than the previous state of the art.

\paragraph{Task-specific Adaptation.} On AGENDA, BART benefits more from our task-adaptive pretraining, achieving the new state of the art of 25.66 BLEU points, a further gain of 2 BLEU points compared to its performance without task adaptation. The improvements from task-adaptive pretraining are not as large as for AMR. We hypothesize that this is due to the fact that the graphs do not completely cover the target text \cite{koncel-kedziorski-etal-2019-text}, making this dataset more challenging. See Table~\ref{tab:results-agenda-appendix} in Appendix for more automatic metrics. 


\begin{table}[t]
\centering
{\renewcommand{\arraystretch}{0.9}
\begin{tabular}{lll} 
\toprule
\textbf{Model} & \multicolumn{2}{c}{\textbf{AMR}}  \\
\midrule
& \, \,\textbf{F} & \,\textbf{MS} \\
\midrule
\citet{mager2020gpttoo} &${5.69}^{{\scriptscriptstyle A}}$&  ${5.08}^{{\scriptscriptstyle A}}$\\
\citet{harkous2020text} &${5.78}^{{\scriptscriptstyle A}}$&  ${5.47}^{{\scriptscriptstyle AB}}$\\
T5\textsubscript{large} &${6.55}^{{\scriptscriptstyle B}}$&  ${6.44}^{{\scriptscriptstyle C}}$\\
BART\textsubscript{large} &${6.70}^{{\scriptscriptstyle B}}$&  ${5.72}^{{\scriptscriptstyle BC}}$\\
Reference &${5.91}^{{\scriptscriptstyle A}}$& - \\
\midrule
\textbf{Model} & \multicolumn{2}{c}{\textbf{WebNLG}}  \\
 \midrule
& \, \,\textbf{F} & \, \,\textbf{SA} \\
\midrule
 \citet{castro-ferreira-etal-2019-neural} &${5.52}^{{\scriptscriptstyle A}}$& ${4.77}^{{\scriptscriptstyle A}}$ \\
 \citet{harkous2020text} &${5.74}^{{\scriptscriptstyle AB}}$&${6.21}^{{\scriptscriptstyle B}}$  \\
 T5\textsubscript{large} &${6.71}^{{\scriptscriptstyle C}}$&  ${6.63}^{{\scriptscriptstyle B}}$\\
BART\textsubscript{large} &${6.53}^{{\scriptscriptstyle C}}$& ${6.50}^{{\scriptscriptstyle B}}$ \\
Reference &${5.89}^{{\scriptscriptstyle B}}$& ${6.47}^{{\scriptscriptstyle B}}$ \\
\bottomrule
\end{tabular}}
\caption{Fluency (F), Meaning Similarity (MS) and Semantic Adequacy (SA) obtained in the human evaluation. Differences between models which have a letter in common are not statistically significant and were determined by pairwise Mann-Whitney tests with $p<0.05$.}
\label{tab:humanevevaluation}
\vspace{-0.3cm}
\end{table}

 \begin{figure*}[t]
    \centering
    \includegraphics[width=1\textwidth]{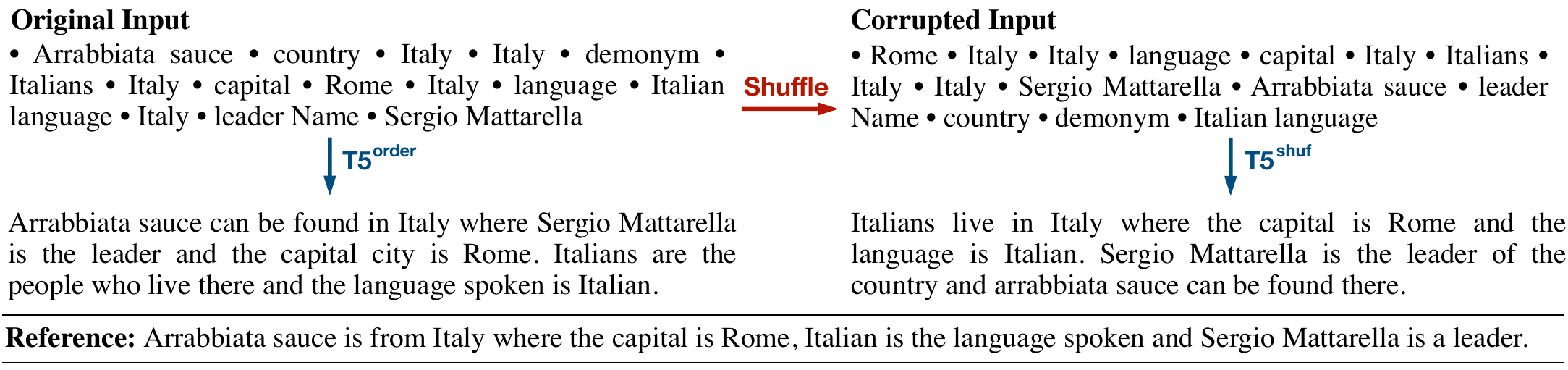}
    \caption{Example graph with 5 triples, from WebNLG dev linearized with the neutral separator tag, denoted \sep{}, (top left), its shuffled version (top right), texts generated with two fine-tuned versions of T5\textsubscript{small} and a gold reference (bottom). Note that T5 can produce a reasonable text even when the input triples are shuffled randomly.}
    \label{fig:graphs-shuffle}
\end{figure*}

 \begin{figure}[t]
    \centering
    \includegraphics[width=.48\textwidth]{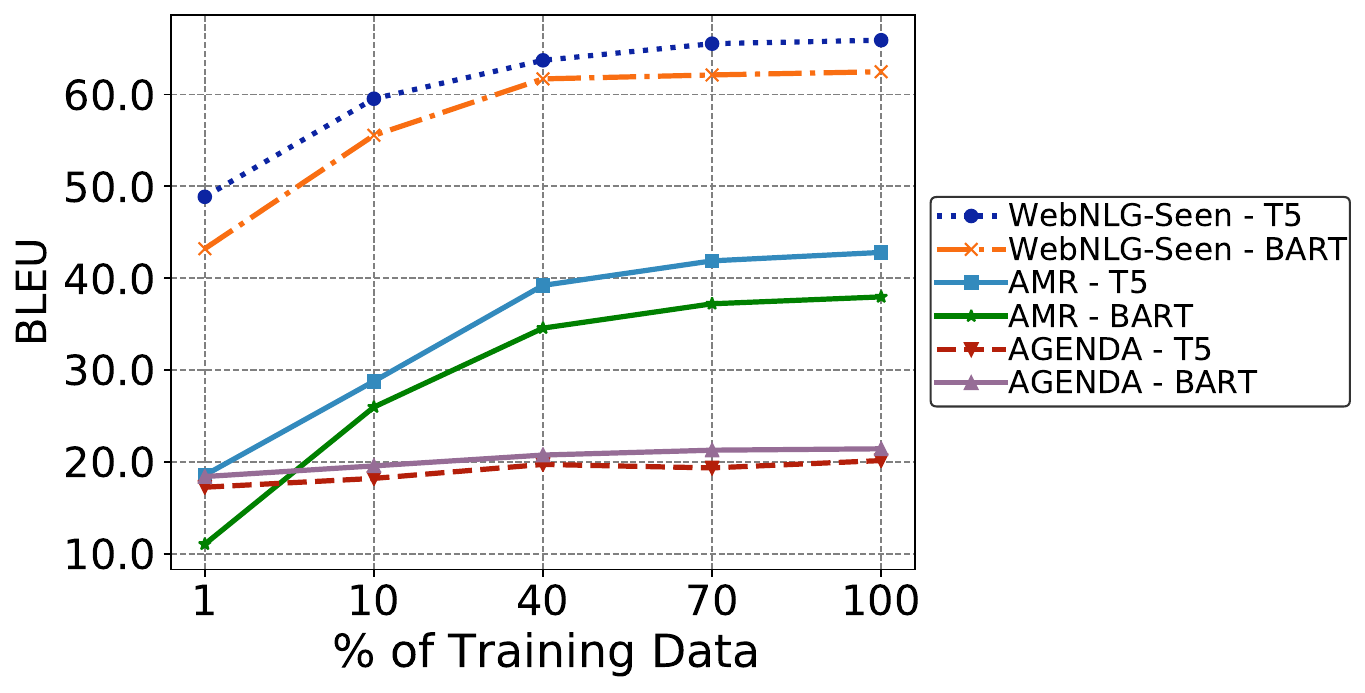}
    \caption{Performance of BART\textsubscript{base} and T5\textsubscript{base} in the dev set when experimenting with different amounts of training data.}
    \label{fig:graphs-trainingexamples}
    \vspace{-3mm}
\end{figure}

\subsection{Human Evaluation}
\label{sec:human_eval}
To further assess the quality of the generated text, we conduct a human evaluation on AMR and WebNLG via crowd sourcing on Amazon Mechanical Turk.\footnote{We exclude AGENDA because its texts are scientific in nature and annotators are not necessarily AI experts.} Following previous works \cite{gardent-etal-2017-webnlg, castro-ferreira-etal-2019-neural}, we assess three quality criteria: (i) \emph{Fluency} (i.e., does the text flow in a natural, easy-to-read manner?), for AMR and WebNLG; (ii) \emph{Meaning Similarity} (i.e., how close in meaning is
the generated text to the reference sentence?) for AMR; (ii) \emph{Semantic Adequacy} (i.e., does the text clearly express the data?) for WebNLG. We randomly select 100 generated texts of each model, which the annotators then rate on a 1-7 Likert scale. For each text, we collect scores from 3 annotators and average them.\footnote{Inter-annotator agreement for the three criteria ranged from 0.40 to 0.79, with an average Krippendorff's $\alpha$ of 0.56.} 

Table~\ref{tab:humanevevaluation} shows the results.  Our approaches improve the fluency, meaning similarity, and semantic adequacy on both datasets compared to other state-of-the-art approaches with statistically significant margins ($p{<}0.05$). Interestingly, the highest fluency improvement ($+0.97$) is on AMR, where our approach also has the largest BLEU improvement ($+8.10$) over \citet{harkous2020text}. Finally, our models score higher than the references in fluency with statistically significant margins, highlighting their strong language generation abilities.\footnote{Examples of fluent generations can be found in the Tables~\ref{tab:human_examples} and \ref{tab:human_examples_agenda} in Appendix.}

\begin{table}[t]
\centering
{\renewcommand{\arraystretch}{0.9}

\begin{tabular}{lrrr}  
\toprule
\textbf{Model} & \textbf{AMR} & \textbf{WebNLG} & \textbf{AGENDA} \\
\midrule

\shufmodel{T5}{small}{order} & 36.83 & 63.41 & 19.86 \\
\shufmodel{T5}{small}{shuf} & 15.56 &  61.54 & 19.08 \\
\bottomrule
\end{tabular}}
\caption{Impact (measured with BLEU) of using a bag of entities and relations (\emph{shuf}) as input for T5\textsubscript{small}.}
\label{tab:shuffle}
\vspace{-3mm}
\end{table}

\subsection{Limiting the Training Data}

In Figure~\ref{fig:graphs-trainingexamples}, we investigate the PLMs' performance, measured with BLEU score, while varying (from 1\% to 100\%) the amount of training data used for fine-tuning. We find that, when fine-tuned with only 40\% of the data, both BART and T5 already greatly improve the performance compared to using the entire training data in all three benchmarks. For example, BART fine-tuned on 40\% of AMR training data achieves 91\% of the BLEU score when fine-tuned on full data.

Note that in a low-resource scenario in AMR and WebNLG, T5 considerably outperforms BART. In particular, with only 1\% of training examples, the difference between T5 and BART is 7.51 and 5.64 BLEU points for AMR and WebNLG, respectively. This suggests that T5 is more data efficient when adapting to the new task, likewise our findings in \textsc{AMR-sta} (cf.\ \S\ref{sec:amr}).

\begin{table*}[t]
    \centering
    \footnotesize
    \begin{tabular}{@{\hspace{.1em}}c@{\hspace{.1em}}c@{\hspace{.1em}}p{4.9cm}@{\hspace{.8em}}p{4.8cm}@{\hspace{.8em}}p{4.75cm}@{\hspace{.2em}}}
    \toprule
         &\textbf{T/F}&\multicolumn{1}{c}{\textbf{Input Fact}} & \multicolumn{1}{c}{\textbf{\shufmodel{T5}{small}{order}}} & \multicolumn{1}{c}{\textbf{\shufmodel{T5}{small}{shuf}}} \\

         \midrule
         (1) &S& \sep{} German language \sep{} Antwerp \sep{} Antwerp \sep{} Antwerp International Airport \sep{} Belgium \sep{} Belgium \sep{} Charles Michel \sep{} city Served \sep{} leader Name \sep{} Belgium \sep{} language \sep{} country & Antwerp International Airport serves the city of Antwerp. German is the language spoken in Belgium where Charles Michel is the leader. & Antwerp International Airport serves the city of Antwerp in Belgium where the German language is spoken and Charles Michel is the leader.\\
        \midrule

          (2) &T& \sep{} California \sep{} is Part Of \sep{} US \sep{} California \sep{} capital \sep{} Sacramento & California is part of the United States and its capital is Sacramento. & California is part of the United States and its capital is Sacramento. \\
         (3) &F& \sep{} US \sep{} is Part Of \sep{} California \sep{} California \sep{} capital \sep{} Sacramento & California's capital is Sacramento and the United States is part of California. & California is part of the United States and its capital is Sacramento.\\
                                   \midrule
            (4) &T& \sep{} Amarillo, Texas \sep{} is Part Of \sep{} United States & Amarillo, Texas is part of the United States. & Amarillo, Texas is part of the United States.\\
         (5) &F& \sep{} United States \sep{} is Part Of \sep{} Amarillo, Texas & Amarillo, Texas is part of the United States. & Amarillo, Texas is part of the United States.   \\

         \bottomrule
    \end{tabular}
    \caption{Example generations from shuffled (S), true (T), and corrupted (F) triple facts by T5\textsubscript{small}, fine-tuned on correctly ordered triples (\emph{order}) and randomly shuffled input (\emph{shuf}).} 
    \label{tab:qualitative}
\end{table*}

\section{Influence of the Graph Structure}

We conduct further experiments to examine how much the PLMs consider the graph structure.
To this end, we remove parentheses in AMRs and replace \ourtag{H}, \ourtag{R}, and \ourtag{T} tokens with neutral separator tokens, denoted \sep{}, for KGs, such that the graph structure is only defined by the order of node and edge labels.
If we shuffle such a sequence,
the graph structure is thus completely obscured and the input effectively becomes a bag of node and edge labels.
See Figure~\ref{fig:graphs-shuffle} for an example of both a correctly ordered and a shuffled triple sequence.

\subsection{Quantitative Analysis}

Table~\ref{tab:shuffle} shows the effect on T5's performance when its input contains correctly ordered triples (\shufmodel{T5}{small}{order}) vs.\ shuffled ones (\shufmodel{T5}{small}{shuf}) for both training and evaluation.
%
%
We first observe that \shufmodel{T5}{small}{order} only has marginally lower performance (around 2-4\%{}) with the neutral separators than with the \ourtag{H}/\ourtag{R}/\ourtag{T} tags or parentheses.\footnote{See a more fine-grained comparison in Appendix~\ref{section:inputgraphsize}.} We see that as evidence that the graph structure is similarly well captured by \shufmodel{T5}{small}{order}.
Without the graph structure (\shufmodel{T5}{small}{shuf}), AMR-to-text performance drops significantly. Possible explanations of this drop are: (i) the relative ordering of the AMR graph is known to correlate with the target sentence order \cite{konsas_17}; (ii) in contrast to WebNLG that contains common knowledge, the AMR dataset contains very specific sentences with higher surprisal;\footnote{Perplexities estimated on the dev sets of AMR and WebNLG datasets, with GPT-2 fine-tuned on the corresponding training set, are 20.9 and 7.8, respectively.} (iii) AMRs are much more complex graph structures than the KGs from WebNLG and AGENDA.\footnote{In Appendix~\ref{sec:graphstats}, we present the graph properties of the datasets and discuss the differences.}  

On the other hand, KG-to-text performance is not much lower,
indicating that most of the PLMs' success in this task stems from their language modeling rather than their graph encoding capabilities. 
We hypothesize that a PLM can match the entities in a shuffled input with sentences mentioning these entities from the pretraining or fine-tuning phase.
It has recently been argued that large PLMs can recall certain common knowledge facts from pretraining \citep{petroni-etal-2019-language,bosselut-etal-2019-comet}.




\subsection{Qualitative Analysis}
\label{sec:qualitative}

The example in Figure~\ref{fig:graphs-shuffle} confirms our impression.
\shufmodel{T5}{small}{shuf} produces a text with the same content as \shufmodel{T5}{small}{order} but does not need the correct triple structure to do so.
Example (1) in Table~\ref{tab:qualitative} shows the output of both models with shuffled input.
Interestingly, even \shufmodel{T5}{small}{order} produces a reasonable and truthful text.
This suggests that previously seen facts serve as a strong guide during text generation, even for models that were fine-tuned with a clearly marked graph structure,
suggesting that \shufmodel{T5}{small}{order} also relies more on language modeling than the graph structure.
It does have more difficulties covering the whole input graph though.
The fact that \emph{Antwerp} is located in \emph{Belgium} is missing from its output.

To further test our hypothesis that
PLMs make use of previously seen facts
during KG-to-text generation, we generate example true facts, corrupt them in a controlled setting, and feed them to both \shufmodel{T5}{small}{order} and \shufmodel{T5}{small}{shuf} to observe their output (examples (2)--(5) in Table~\ref{tab:qualitative}).
The model trained on correctly ordered input has learned a bit more to rely on the input graph structure. 
The false fact in example (3) with two triples is reliably transferred to the text by \shufmodel{T5}{small}{order} but not by \shufmodel{T5}{small}{shuf}, which silently corrects it.
Also note that, in example (5), both models refuse to generate an incorrect fact. More examples can be found in Table~\ref{tab:qualitative_appendix} in the Appendix.

Our qualitative analysis illustrates that state-of-the-art PLMs, despite their fluency capacities (cf.\ \S\ref{sec:human_eval}), bear the risk of parroting back training sentences while ignoring the input structure. This issue can limit the practical usage of those models as, in many cases, it is important for a generation model to stay true to its input \cite{wiseman-etal-2017-challenges, falke-etal-2019-ranking}.





\section{Conclusion}

We investigated two pretrained language models (PLMs) for graph-to-text generation and show that the pretraining strategies, language model adaptation (\textsc{lma}) and supervised task adaptation (\textsc{sta}), can lead to notable improvements. Our approaches outperform the state of the art by a substantial margin on three graph-to-text benchmarks. Moreover, in a human evaluation our generated texts are perceived significantly more fluent than human references. Examining the influence of the graph structure on the text generation process, we find that PLMs may not always follow the graph structure and instead use memorized facts to guide the generation. A promising direction for future work is to explore ways of injecting a stronger graph-structural bias into PLMs, thus possibly leveraging their strong language modeling capabilities and keeping the output faithful to the input graph.

\section*{Acknowledgments}
We thank our anonymous reviewers for their thoughtful feedback. Leonardo F. R. Ribeiro is supported by the German Research Foundation (DFG) as part of the Research Training Group ``Adaptive Preparation of Information form Heterogeneous Sources'' (AIPHES, GRK 1994/1) and as part of the DFG funded project UKP-SQuARE with the number GU 798/29-1. Martin Schmitt is supported by the BMBF as part of the project MLWin (01IS18050) and by the German Academic Scholarship Foundation (Studienstiftung des deutschen Volkes).

\bibliography{anthology,custom}

\begin{thebibliography}{67}
\expandafter\ifx\csname natexlab\endcsname\relax\def\natexlab#1{#1}\fi

\bibitem[{Ammar et~al.(2018)Ammar, Groeneveld, Bhagavatula, Beltagy, Crawford,
  Downey, Dunkelberger, Elgohary, Feldman, Ha, Kinney, Kohlmeier, Lo, Murray,
  Ooi, Peters, Power, Skjonsberg, Wang, Wilhelm, Yuan, van Zuylen, and
  Etzioni}]{ammar-etal-2018-construction}
Waleed Ammar, Dirk Groeneveld, Chandra Bhagavatula, Iz~Beltagy, Miles Crawford,
  Doug Downey, Jason Dunkelberger, Ahmed Elgohary, Sergey Feldman, Vu~Ha,
  Rodney Kinney, Sebastian Kohlmeier, Kyle Lo, Tyler Murray, Hsu-Han Ooi,
  Matthew Peters, Joanna Power, Sam Skjonsberg, Lucy Wang, Chris Wilhelm, Zheng
  Yuan, Madeleine van Zuylen, and Oren Etzioni. 2018.
\newblock \href {https://doi.org/10.18653/v1/N18-3011} {Construction of the
  literature graph in semantic scholar}.
\newblock In \emph{Proceedings of the 2018 Conference of the North {A}merican
  Chapter of the Association for Computational Linguistics: Human Language
  Technologies, Volume 3 (Industry Papers)}, pages 84--91, New Orleans -
  Louisiana. Association for Computational Linguistics.

\bibitem[{An(2019)}]{An2019RepulsiveBS}
Bang An. 2019.
\newblock \href {http://bayesiandeeplearning.org/2019/papers/25.pdf} {Repulsive
  bayesian sampling for diversified attention modeling}.
\newblock In \emph{4th workshop on Bayesian Deep Learning (NeurIPS 2019)}.

\bibitem[{Auer et~al.(2007)Auer, Bizer, Kobilarov, Lehmann, Cyganiak, and
  Ives}]{10.5555/1785162.1785216}
S\"{o}ren Auer, Christian Bizer, Georgi Kobilarov, Jens Lehmann, Richard
  Cyganiak, and Zachary Ives. 2007.
\newblock \href
  {https://link.springer.com/chapter/10.1007/978-3-540-76298-0_52} {Dbpedia: A
  nucleus for a web of open data}.
\newblock In \emph{Proceedings of the 6th International The Semantic Web and
  2nd Asian Conference on Asian Semantic Web Conference}, ISWC’07/ASWC’07,
  page 722–735, Berlin, Heidelberg. Springer-Verlag.

\bibitem[{Bai et~al.(2021)Bai, Chen, Song, and Zhang}]{bai-etal-2021-semantic}
Xuefeng Bai, Yulong Chen, Linfeng Song, and Yue Zhang. 2021.
\newblock \href {https://doi.org/10.18653/v1/2021.acl-long.342} {Semantic
  representation for dialogue modeling}.
\newblock In \emph{Proceedings of the 59th Annual Meeting of the Association
  for Computational Linguistics and the 11th International Joint Conference on
  Natural Language Processing (Volume 1: Long Papers)}, pages 4430--4445,
  Online. Association for Computational Linguistics.

\bibitem[{Banarescu et~al.(2013)Banarescu, Bonial, Cai, Georgescu, Griffitt,
  Hermjakob, Knight, Koehn, Palmer, and
  Schneider}]{banarescu-etal-2013-abstract}
Laura Banarescu, Claire Bonial, Shu Cai, Madalina Georgescu, Kira Griffitt, Ulf
  Hermjakob, Kevin Knight, Philipp Koehn, Martha Palmer, and Nathan Schneider.
  2013.
\newblock \href {https://www.aclweb.org/anthology/W13-2322} {{A}bstract
  {M}eaning {R}epresentation for sembanking}.
\newblock In \emph{Proceedings of the 7th Linguistic Annotation Workshop and
  Interoperability with Discourse}, pages 178--186, Sofia, Bulgaria.
  Association for Computational Linguistics.

\bibitem[{Beck et~al.(2018)Beck, Haffari, and Cohn}]{beck-etal-2018-acl2018}
Daniel Beck, Gholamreza Haffari, and Trevor Cohn. 2018.
\newblock \href {https://www.aclweb.org/anthology/P18-1026} {Graph-to-sequence
  learning using gated graph neural networks}.
\newblock In \emph{Proceedings of the 56th Annual Meeting of the Association
  for Computational Linguistics (Volume 1: Long Papers)}, pages 273--283,
  Melbourne, Australia. Association for Computational Linguistics.

\bibitem[{Belz et~al.(2011)Belz, White, Espinosa, Kow, Hogan, and
  Stent}]{belz-etal-2011-first}
Anja Belz, Michael White, Dominic Espinosa, Eric Kow, Deirdre Hogan, and Amanda
  Stent. 2011.
\newblock \href {https://www.aclweb.org/anthology/W11-2832} {The first surface
  realisation shared task: Overview and evaluation results}.
\newblock In \emph{Proceedings of the 13th {E}uropean Workshop on Natural
  Language Generation}, pages 217--226, Nancy, France. Association for
  Computational Linguistics.

\bibitem[{Bonial et~al.(2020)Bonial, Donatelli, Abrams, Lukin, Tratz, Marge,
  Artstein, Traum, and Voss}]{bonial-etal-2020-dialogue}
Claire Bonial, Lucia Donatelli, Mitchell Abrams, Stephanie~M. Lukin, Stephen
  Tratz, Matthew Marge, Ron Artstein, David Traum, and Clare Voss. 2020.
\newblock \href {https://www.aclweb.org/anthology/2020.lrec-1.86}
  {Dialogue-{AMR}: {A}bstract {M}eaning {R}epresentation for dialogue}.
\newblock In \emph{Proceedings of the 12th Language Resources and Evaluation
  Conference}, pages 684--695, Marseille, France. European Language Resources
  Association.

\bibitem[{Bosselut et~al.(2019)Bosselut, Rashkin, Sap, Malaviya, Celikyilmaz,
  and Choi}]{bosselut-etal-2019-comet}
Antoine Bosselut, Hannah Rashkin, Maarten Sap, Chaitanya Malaviya, Asli
  Celikyilmaz, and Yejin Choi. 2019.
\newblock \href {https://doi.org/10.18653/v1/P19-1470} {{COMET}: Commonsense
  transformers for automatic knowledge graph construction}.
\newblock In \emph{Proceedings of the 57th Annual Meeting of the Association
  for Computational Linguistics}, pages 4762--4779, Florence, Italy.
  Association for Computational Linguistics.

\bibitem[{Cai and Lam(2020{\natexlab{a}})}]{cai-lam-2020-amr}
Deng Cai and Wai Lam. 2020{\natexlab{a}}.
\newblock \href {https://doi.org/10.18653/v1/2020.acl-main.119} {{AMR} parsing
  via graph-sequence iterative inference}.
\newblock In \emph{Proceedings of the 58th Annual Meeting of the Association
  for Computational Linguistics}, pages 1290--1301, Online. Association for
  Computational Linguistics.

\bibitem[{Cai and Lam(2020{\natexlab{b}})}]{cai-lam-2020-graph}
Deng Cai and Wai Lam. 2020{\natexlab{b}}.
\newblock \href {https://aaai.org/ojs/index.php/AAAI/article/view/6243} {Graph
  transformer for graph-to-sequence learning}.
\newblock In \emph{The Thirty-Fourth {AAAI} Conference on Artificial
  Intelligence, {AAAI} 2020, The Thirty-Second Innovative Applications of
  Artificial Intelligence Conference, {IAAI} 2020, The Tenth {AAAI} Symposium
  on Educational Advances in Artificial Intelligence, {EAAI} 2020, New York,
  NY, USA, February 7-12, 2020}, pages 7464--7471. {AAAI} Press.

\bibitem[{Castro~Ferreira et~al.(2019)Castro~Ferreira, van~der Lee, van
  Miltenburg, and Krahmer}]{castro-ferreira-etal-2019-neural}
Thiago Castro~Ferreira, Chris van~der Lee, Emiel van Miltenburg, and Emiel
  Krahmer. 2019.
\newblock \href {https://doi.org/10.18653/v1/D19-1052} {Neural data-to-text
  generation: A comparison between pipeline and end-to-end architectures}.
\newblock In \emph{Proceedings of the 2019 Conference on Empirical Methods in
  Natural Language Processing and the 9th International Joint Conference on
  Natural Language Processing (EMNLP-IJCNLP)}, pages 552--562, Hong Kong,
  China. Association for Computational Linguistics.

\bibitem[{Cheng et~al.(2020)Cheng, Agrawal, Mart{\'\i}nez~Alonso, Bhargava,
  Driesen, Flego, Kaplan, Kartsaklis, Li, Piraviperumal, Williams, Yu,
  {\'O}~S{\'e}aghdha, and Johannsen}]{cheng-etal-2020-conversational}
Jianpeng Cheng, Devang Agrawal, H{\'e}ctor Mart{\'\i}nez~Alonso, Shruti
  Bhargava, Joris Driesen, Federico Flego, Dain Kaplan, Dimitri Kartsaklis, Lin
  Li, Dhivya Piraviperumal, Jason~D. Williams, Hong Yu, Diarmuid
  {\'O}~S{\'e}aghdha, and Anders Johannsen. 2020.
\newblock \href {https://doi.org/10.18653/v1/2020.emnlp-main.651}
  {Conversational semantic parsing for dialog state tracking}.
\newblock In \emph{Proceedings of the 2020 Conference on Empirical Methods in
  Natural Language Processing (EMNLP)}, pages 8107--8117, Online. Association
  for Computational Linguistics.

\bibitem[{Damonte and Cohen(2019)}]{damonte-cohen-2019-structural}
Marco Damonte and Shay~B. Cohen. 2019.
\newblock \href {https://doi.org/10.18653/v1/N19-1366} {Structural neural
  encoders for {AMR}-to-text generation}.
\newblock In \emph{Proceedings of the 2019 Conference of the North {A}merican
  Chapter of the Association for Computational Linguistics: Human Language
  Technologies, Volume 1 (Long and Short Papers)}, pages 3649--3658,
  Minneapolis, Minnesota. Association for Computational Linguistics.

\bibitem[{Denkowski and Lavie(2014)}]{Denkowski14meteoruniversal}
Michael Denkowski and Alon Lavie. 2014.
\newblock \href {https://doi.org/10.3115/v1/W14-3348} {Meteor universal:
  Language specific translation evaluation for any target language}.
\newblock In \emph{Proceedings of the Ninth Workshop on Statistical Machine
  Translation}, pages 376--380, Baltimore, Maryland, USA. Association for
  Computational Linguistics.

\bibitem[{Devlin et~al.(2019)Devlin, Chang, Lee, and
  Toutanova}]{devlin-etal-2019-bert}
Jacob Devlin, Ming-Wei Chang, Kenton Lee, and Kristina Toutanova. 2019.
\newblock \href {https://doi.org/10.18653/v1/N19-1423} {{BERT}: Pre-training of
  deep bidirectional transformers for language understanding}.
\newblock In \emph{Proceedings of the 2019 Conference of the North {A}merican
  Chapter of the Association for Computational Linguistics: Human Language
  Technologies, Volume 1 (Long and Short Papers)}, pages 4171--4186,
  Minneapolis, Minnesota. Association for Computational Linguistics.

\bibitem[{Duan et~al.(2017)Duan, Tang, Chen, and
  Zhou}]{duan-etal-2017-question}
Nan Duan, Duyu Tang, Peng Chen, and Ming Zhou. 2017.
\newblock \href {https://doi.org/10.18653/v1/D17-1090} {Question generation for
  question answering}.
\newblock In \emph{Proceedings of the 2017 Conference on Empirical Methods in
  Natural Language Processing}, pages 866--874, Copenhagen, Denmark.
  Association for Computational Linguistics.

\bibitem[{Falke et~al.(2019)Falke, Ribeiro, Utama, Dagan, and
  Gurevych}]{falke-etal-2019-ranking}
Tobias Falke, Leonardo F.~R. Ribeiro, Prasetya~Ajie Utama, Ido Dagan, and Iryna
  Gurevych. 2019.
\newblock \href {https://doi.org/10.18653/v1/P19-1213} {Ranking generated
  summaries by correctness: An interesting but challenging application for
  natural language inference}.
\newblock In \emph{Proceedings of the 57th Annual Meeting of the Association
  for Computational Linguistics}, pages 2214--2220, Florence, Italy.
  Association for Computational Linguistics.

\bibitem[{Gardent et~al.(2017)Gardent, Shimorina, Narayan, and
  Perez-Beltrachini}]{gardent-etal-2017-webnlg}
Claire Gardent, Anastasia Shimorina, Shashi Narayan, and Laura
  Perez-Beltrachini. 2017.
\newblock \href {https://doi.org/10.18653/v1/W17-3518} {The {W}eb{NLG}
  challenge: Generating text from {RDF} data}.
\newblock In \emph{Proceedings of the 10th International Conference on Natural
  Language Generation}, pages 124--133, Santiago de Compostela, Spain.
  Association for Computational Linguistics.

\bibitem[{Gatt and Krahmer(2018)}]{10.5555/3241691.3241693}
Albert Gatt and Emiel Krahmer. 2018.
\newblock \href {https://dl.acm.org/doi/10.5555/3241691.3241693} {Survey of the
  state of the art in natural language generation: Core tasks, applications and
  evaluation}.
\newblock \emph{Journal of Artificial Intelligence Research}, 61(1):65–170.

\bibitem[{Guo et~al.(2019)Guo, Zhang, Teng, and Lu}]{dcgcnforgraph2seq19guo}
Zhijiang Guo, Yan Zhang, Zhiyang Teng, and Wei Lu. 2019.
\newblock \href {https://doi.org/10.1162/tacl_a_00269} {Densely connected graph
  convolutional networks for graph-to-sequence learning}.
\newblock \emph{Transactions of the Association for Computational Linguistics},
  7:297--312.

\bibitem[{Gururangan et~al.(2020)Gururangan, Marasovi{\'c}, Swayamdipta, Lo,
  Beltagy, Downey, and Smith}]{gururangan-etal-2020-dont}
Suchin Gururangan, Ana Marasovi{\'c}, Swabha Swayamdipta, Kyle Lo, Iz~Beltagy,
  Doug Downey, and Noah~A. Smith. 2020.
\newblock \href {https://doi.org/10.18653/v1/2020.acl-main.740} {Don{'}t stop
  pretraining: Adapt language models to domains and tasks}.
\newblock In \emph{Proceedings of the 58th Annual Meeting of the Association
  for Computational Linguistics}, pages 8342--8360, Online. Association for
  Computational Linguistics.

\bibitem[{Harkous et~al.(2020)Harkous, Groves, and Saffari}]{harkous2020text}
Hamza Harkous, Isabel Groves, and Amir Saffari. 2020.
\newblock \href {https://doi.org/10.18653/v1/2020.coling-main.218} {Have your
  text and use it too! end-to-end neural data-to-text generation with semantic
  fidelity}.
\newblock In \emph{Proceedings of the 28th International Conference on
  Computational Linguistics}, pages 2410--2424, Barcelona, Spain (Online).
  International Committee on Computational Linguistics.

\bibitem[{Hoyle et~al.(2021)Hoyle, Marasovi{\'c}, and
  Smith}]{hoyle2020promoting}
Alexander~Miserlis Hoyle, Ana Marasovi{\'c}, and Noah~A. Smith. 2021.
\newblock \href {https://doi.org/10.18653/v1/2021.findings-acl.82} {Promoting
  graph awareness in linearized graph-to-text generation}.
\newblock In \emph{Findings of the Association for Computational Linguistics:
  ACL-IJCNLP 2021}, pages 944--956, Online. Association for Computational
  Linguistics.

\bibitem[{Kale(2020)}]{kale2020texttotext}
Mihir Kale. 2020.
\newblock \href {http://arxiv.org/abs/2005.10433} {Text-to-text pre-training
  for data-to-text tasks}.
\newblock \emph{arXiv e-prints}.

\bibitem[{Kingma and Ba(2015)}]{kingma:adam}
Diederik~P. Kingma and Jimmy Ba. 2015.
\newblock \href {http://arxiv.org/abs/1412.6980} {Adam: {A} method for
  stochastic optimization}.
\newblock In \emph{3rd International Conference on Learning Representations,
  {ICLR} 2015, San Diego, CA, USA, May 7-9, 2015, Conference Track
  Proceedings}.

\bibitem[{Kipf and Welling(2017)}]{Kipf:2016tc}
Thomas~N. Kipf and Max Welling. 2017.
\newblock \href {https://openreview.net/forum?id=SJU4ayYgl} {{Semi-Supervised
  Classification with Graph Convolutional Networks}}.
\newblock In \emph{Proceedings of the 5th International Conference on Learning
  Representations}, ICLR 2017.

\bibitem[{Koncel-Kedziorski et~al.(2019)Koncel-Kedziorski, Bekal, Luan, Lapata,
  and Hajishirzi}]{koncel-kedziorski-etal-2019-text}
Rik Koncel-Kedziorski, Dhanush Bekal, Yi~Luan, Mirella Lapata, and Hannaneh
  Hajishirzi. 2019.
\newblock \href {https://doi.org/10.18653/v1/N19-1238} {{T}ext {G}eneration
  from {K}nowledge {G}raphs with {G}raph {T}ransformers}.
\newblock In \emph{Proceedings of the 2019 Conference of the North {A}merican
  Chapter of the Association for Computational Linguistics: Human Language
  Technologies, Volume 1 (Long and Short Papers)}, pages 2284--2293,
  Minneapolis, Minnesota. Association for Computational Linguistics.

\bibitem[{Konstas et~al.(2017)Konstas, Iyer, Yatskar, Choi, and
  Zettlemoyer}]{konsas_17}
Ioannis Konstas, Srinivasan Iyer, Mark Yatskar, Yejin Choi, and Luke
  Zettlemoyer. 2017.
\newblock \href {https://doi.org/10.18653/v1/P17-1014} {Neural amr:
  Sequence-to-sequence models for parsing and generation}.
\newblock In \emph{Proceedings of the 55th Annual Meeting of the Association
  for Computational Linguistics (Volume 1: Long Papers)}, pages 146--157,
  Vancouver, Canada. Association for Computational Linguistics.

\bibitem[{Lewis et~al.(2020)Lewis, Liu, Goyal, Ghazvininejad, Mohamed, Levy,
  Stoyanov, and Zettlemoyer}]{lewis2019bart}
Mike Lewis, Yinhan Liu, Naman Goyal, Marjan Ghazvininejad, Abdelrahman Mohamed,
  Omer Levy, Veselin Stoyanov, and Luke Zettlemoyer. 2020.
\newblock \href {https://www.aclweb.org/anthology/2020.acl-main.703} {{BART}:
  Denoising sequence-to-sequence pre-training for natural language generation,
  translation, and comprehension}.
\newblock In \emph{Proceedings of the 58th Annual Meeting of the Association
  for Computational Linguistics}, pages 7871--7880, Online. Association for
  Computational Linguistics.

\bibitem[{Liu et~al.(2020)Liu, Ott, Goyal, Du, Joshi, Chen, Levy, Lewis,
  Zettlemoyer, and Stoyanov}]{liu2020roberta}
Yinhan Liu, Myle Ott, Naman Goyal, Jingfei Du, Mandar Joshi, Danqi Chen, Omer
  Levy, Mike Lewis, Luke Zettlemoyer, and Veselin Stoyanov. 2020.
\newblock \href {https://openreview.net/forum?id=SyxS0T4tvS} {Roberta: A
  robustly optimized bert pretraining approach}.
\newblock \emph{arXiv e-prints}.

\bibitem[{Luan et~al.(2018)Luan, He, Ostendorf, and
  Hajishirzi}]{luan-etal-2018-multi}
Yi~Luan, Luheng He, Mari Ostendorf, and Hannaneh Hajishirzi. 2018.
\newblock \href {https://doi.org/10.18653/v1/D18-1360} {Multi-task
  identification of entities, relations, and coreference for scientific
  knowledge graph construction}.
\newblock In \emph{Proceedings of the 2018 Conference on Empirical Methods in
  Natural Language Processing}, pages 3219--3232, Brussels, Belgium.
  Association for Computational Linguistics.

\bibitem[{Mager et~al.(2020)Mager, Fernandez~Astudillo, Naseem, Sultan, Lee,
  Florian, and Roukos}]{mager2020gpttoo}
Manuel Mager, Ram{\'o}n Fernandez~Astudillo, Tahira Naseem, Md~Arafat Sultan,
  Young-Suk Lee, Radu Florian, and Salim Roukos. 2020.
\newblock \href {https://www.aclweb.org/anthology/2020.acl-main.167}
  {{GPT}-too: A language-model-first approach for {AMR}-to-text generation}.
\newblock In \emph{Proceedings of the 58th Annual Meeting of the Association
  for Computational Linguistics}, pages 1846--1852, Online. Association for
  Computational Linguistics.

\bibitem[{Marcheggiani and Perez~Beltrachini(2018)}]{marcheggiani-icnl18}
Diego Marcheggiani and Laura Perez~Beltrachini. 2018.
\newblock \href {https://www.aclweb.org/anthology/W18-6501} {Deep graph
  convolutional encoders for structured data to text generation}.
\newblock In \emph{Proceedings of the 11th International Conference on Natural
  Language Generation}, pages 1--9, Tilburg University, The Netherlands.
  Association for Computational Linguistics.

\bibitem[{Moon et~al.(2019)Moon, Shah, Kumar, and
  Subba}]{moon-etal-2019-opendialkg}
Seungwhan Moon, Pararth Shah, Anuj Kumar, and Rajen Subba. 2019.
\newblock \href {https://doi.org/10.18653/v1/P19-1081} {{O}pen{D}ial{KG}:
  Explainable conversational reasoning with attention-based walks over
  knowledge graphs}.
\newblock In \emph{Proceedings of the 57th Annual Meeting of the Association
  for Computational Linguistics}, pages 845--854, Florence, Italy. Association
  for Computational Linguistics.

\bibitem[{Moryossef et~al.(2019)Moryossef, Goldberg, and
  Dagan}]{moryossef-etal-2019-step}
Amit Moryossef, Yoav Goldberg, and Ido Dagan. 2019.
\newblock \href {https://doi.org/10.18653/v1/N19-1236} {{S}tep-by-step:
  {S}eparating planning from realization in neural data-to-text generation}.
\newblock In \emph{Proceedings of the 2019 Conference of the North {A}merican
  Chapter of the Association for Computational Linguistics: Human Language
  Technologies, Volume 1 (Long and Short Papers)}, pages 2267--2277,
  Minneapolis, Minnesota. Association for Computational Linguistics.

\bibitem[{Papineni et~al.(2002)Papineni, Roukos, Ward, and
  Zhu}]{Papineni:2002:BMA:1073083.1073135}
Kishore Papineni, Salim Roukos, Todd Ward, and Wei-Jing Zhu. 2002.
\newblock \href {https://doi.org/10.3115/1073083.1073135} {Bleu: A method for
  automatic evaluation of machine translation}.
\newblock In \emph{Proceedings of the 40th Annual Meeting on Association for
  Computational Linguistics}, ACL '02, pages 311--318, Stroudsburg, PA, USA.
  Association for Computational Linguistics.

\bibitem[{Petroni et~al.(2019)Petroni, Rockt{\"a}schel, Riedel, Lewis, Bakhtin,
  Wu, and Miller}]{petroni-etal-2019-language}
Fabio Petroni, Tim Rockt{\"a}schel, Sebastian Riedel, Patrick Lewis, Anton
  Bakhtin, Yuxiang Wu, and Alexander Miller. 2019.
\newblock \href {https://doi.org/10.18653/v1/D19-1250} {Language models as
  knowledge bases?}
\newblock In \emph{Proceedings of the 2019 Conference on Empirical Methods in
  Natural Language Processing and the 9th International Joint Conference on
  Natural Language Processing (EMNLP-IJCNLP)}, pages 2463--2473, Hong Kong,
  China. Association for Computational Linguistics.

\bibitem[{Popovi{\'c}(2015)}]{popovic-2015-chrf}
Maja Popovi{\'c}. 2015.
\newblock \href {https://doi.org/10.18653/v1/W15-3049} {chr{F}: character
  n-gram {F}-score for automatic {MT} evaluation}.
\newblock In \emph{Proceedings of the Tenth Workshop on Statistical Machine
  Translation}, pages 392--395, Lisbon, Portugal. Association for Computational
  Linguistics.

\bibitem[{Radev et~al.(2020)Radev, Zhang, Rau, Sivaprasad, Hsieh, Rajani, Tang,
  Vyas, Verma, Krishna, Liu, Irwanto, Pan, Rahman, Zaidi, Mutuma, Tarabar,
  Gupta, Yu, Tan, Lin, Xiong, and Socher}]{radev2020dart}
Dragomir Radev, Rui Zhang, Amrit Rau, Abhinand Sivaprasad, Chiachun Hsieh,
  Nazneen~Fatema Rajani, Xiangru Tang, Aadit Vyas, Neha Verma, Pranav Krishna,
  Yangxiaokang Liu, Nadia Irwanto, Jessica Pan, Faiaz Rahman, Ahmad Zaidi,
  Murori Mutuma, Yasin Tarabar, Ankit Gupta, Tao Yu, Yi~Chern Tan, Xi~Victoria
  Lin, Caiming Xiong, and Richard Socher. 2020.
\newblock \href {http://arxiv.org/abs/2007.02871} {Dart: Open-domain structured
  data record to text generation}.
\newblock \emph{arXiv e-prints}.

\bibitem[{Radford et~al.(2019)Radford, Wu, Child, Luan, Amodei, and
  Sutskever}]{radford2019language}
Alec Radford, Jeffrey Wu, Rewon Child, David Luan, Dario Amodei, and Ilya
  Sutskever. 2019.
\newblock \href
  {https://d4mucfpksywv.cloudfront.net/better-language-models/language-models.pdf}
  {Language models are unsupervised multitask learners}.
\newblock \emph{arXiv e-prints}.

\bibitem[{Raffel et~al.(2019)Raffel, Shazeer, Roberts, Lee, Narang, Matena,
  Zhou, Li, and Liu}]{2019t5}
Colin Raffel, Noam Shazeer, Adam Roberts, Katherine Lee, Sharan Narang, Michael
  Matena, Yanqi Zhou, Wei Li, and Peter~J. Liu. 2019.
\newblock \href {http://arxiv.org/abs/1910.10683} {Exploring the limits of
  transfer learning with a unified text-to-text transformer}.
\newblock \emph{arXiv e-prints}.

\bibitem[{Ribeiro et~al.(2019)Ribeiro, Gardent, and
  Gurevych}]{ribeiro-etal-2019-enhancing}
Leonardo F.~R. Ribeiro, Claire Gardent, and Iryna Gurevych. 2019.
\newblock \href {https://doi.org/10.18653/v1/D19-1314} {Enhancing {AMR}-to-text
  generation with dual graph representations}.
\newblock In \emph{Proceedings of the 2019 Conference on Empirical Methods in
  Natural Language Processing and the 9th International Joint Conference on
  Natural Language Processing (EMNLP-IJCNLP)}, pages 3183--3194, Hong Kong,
  China. Association for Computational Linguistics.

\bibitem[{Ribeiro et~al.(2021{\natexlab{a}})Ribeiro, Pfeiffer, Zhang, and
  Gurevych}]{ribeiro2021smelting}
Leonardo F.~R. Ribeiro, Jonas Pfeiffer, Yue Zhang, and Iryna Gurevych.
  2021{\natexlab{a}}.
\newblock \href {https://arxiv.org/pdf/2109.03808.pdf} {Smelting gold and
  silver for improved multilingual amr-to-text generation}.
\newblock In \emph{Proceedings of the 2021 Conference on Empirical Methods in
  Natural Language Processing, {EMNLP} 2021, Punta Cana, November 7-11, 2021}.

\bibitem[{Ribeiro et~al.(2020)Ribeiro, Zhang, Gardent, and
  Gurevych}]{ribeiro-etal-2020-modeling}
Leonardo F.~R. Ribeiro, Yue Zhang, Claire Gardent, and Iryna Gurevych. 2020.
\newblock \href {https://doi.org/10.1162/tacl_a_00332} {Modeling global and
  local node contexts for text generation from knowledge graphs}.
\newblock \emph{Transactions of the Association for Computational Linguistics},
  8:589--604.

\bibitem[{Ribeiro et~al.(2021{\natexlab{b}})Ribeiro, Zhang, and
  Gurevych}]{ribeiro2021structural}
Leonardo F.~R. Ribeiro, Yue Zhang, and Iryna Gurevych. 2021{\natexlab{b}}.
\newblock \href {http://arxiv.org/abs/2103.09120} {Structural adapters in
  pretrained language models for amr-to-text generation}.
\newblock In \emph{Proceedings of the 2021 Conference on Empirical Methods in
  Natural Language Processing, {EMNLP} 2021, Punta Cana, November 7-11, 2021}.

\bibitem[{Schmitt et~al.(2021)Schmitt, Ribeiro, Dufter, Gurevych, and
  Sch{\"u}tze}]{schmitt-etal-2021-modeling}
Martin Schmitt, Leonardo F.~R. Ribeiro, Philipp Dufter, Iryna Gurevych, and
  Hinrich Sch{\"u}tze. 2021.
\newblock \href {https://aclanthology.org/2021.textgraphs-1.2} {Modeling graph
  structure via relative position for text generation from knowledge graphs}.
\newblock In \emph{Proceedings of the Fifteenth Workshop on Graph-Based Methods
  for Natural Language Processing (TextGraphs-15)}, pages 10--21, Mexico City,
  Mexico. Association for Computational Linguistics.

\bibitem[{Schmitt et~al.(2020)Schmitt, Ribeiro, Dufter, Gurevych, and
  Schütze}]{schmitt2020modeling}
Martin Schmitt, Leonardo F.~R. Ribeiro, Philipp Dufter, Iryna Gurevych, and
  Hinrich Schütze. 2020.
\newblock \href {http://arxiv.org/abs/2006.09242} {Modeling graph structure via
  relative position for better text generation from knowledge graphs}.
\newblock \emph{arXiv e-prints}.

\bibitem[{Sellam et~al.(2020)Sellam, Das, and Parikh}]{sellam-etal-2020-bleurt}
Thibault Sellam, Dipanjan Das, and Ankur Parikh. 2020.
\newblock \href {https://doi.org/10.18653/v1/2020.acl-main.704} {{BLEURT}:
  Learning robust metrics for text generation}.
\newblock In \emph{Proceedings of the 58th Annual Meeting of the Association
  for Computational Linguistics}, pages 7881--7892, Online. Association for
  Computational Linguistics.

\bibitem[{Song et~al.(2018)Song, Zhang, Wang, and Gildea}]{song-etal-acl2018}
Linfeng Song, Yue Zhang, Zhiguo Wang, and Daniel Gildea. 2018.
\newblock \href {https://www.aclweb.org/anthology/P18-1150} {A
  graph-to-sequence model for {AMR}-to-text generation}.
\newblock In \emph{Proceedings of the 56th Annual Meeting of the Association
  for Computational Linguistics (Volume 1: Long Papers)}, pages 1616--1626,
  Melbourne, Australia. Association for Computational Linguistics.

\bibitem[{Surdeanu et~al.(2008)Surdeanu, Johansson, Meyers, M{\`a}rquez, and
  Nivre}]{surdeanu-etal-2008-conll}
Mihai Surdeanu, Richard Johansson, Adam Meyers, Llu{\'\i}s M{\`a}rquez, and
  Joakim Nivre. 2008.
\newblock \href {https://www.aclweb.org/anthology/W08-2121} {The {C}o{NLL} 2008
  shared task on joint parsing of syntactic and semantic dependencies}.
\newblock In \emph{{C}o{NLL} 2008: Proceedings of the Twelfth Conference on
  Computational Natural Language Learning}, pages 159--177, Manchester,
  England. Coling 2008 Organizing Committee.

\bibitem[{Trisedya et~al.(2018)Trisedya, Qi, Zhang, and
  Wang}]{trisedya-etal-2018-gtr}
Bayu~Distiawan Trisedya, Jianzhong Qi, Rui Zhang, and Wei Wang. 2018.
\newblock \href {https://doi.org/10.18653/v1/P18-1151} {{GTR}-{LSTM}: A triple
  encoder for sentence generation from {RDF} data}.
\newblock In \emph{Proceedings of the 56th Annual Meeting of the Association
  for Computational Linguistics (Volume 1: Long Papers)}, pages 1627--1637,
  Melbourne, Australia. Association for Computational Linguistics.

\bibitem[{Vaswani et~al.(2017)Vaswani, Shazeer, Parmar, Uszkoreit, Jones,
  Gomez, Kaiser, and Polosukhin}]{NIPS2017_7181}
Ashish Vaswani, Noam Shazeer, Niki Parmar, Jakob Uszkoreit, Llion Jones,
  Aidan~N Gomez, \L~ukasz Kaiser, and Illia Polosukhin. 2017.
\newblock \href
  {http://papers.nips.cc/paper/7181-attention-is-all-you-need.pdf} {Attention
  is all you need}.
\newblock In I.~Guyon, U.~V. Luxburg, S.~Bengio, H.~Wallach, R.~Fergus,
  S.~Vishwanathan, and R.~Garnett, editors, \emph{Advances in Neural
  Information Processing Systems 30}, pages 5998--6008. Curran Associates, Inc.

\bibitem[{Vougiouklis et~al.(2018)Vougiouklis, Elsahar, Kaffee, Gravier,
  Laforest, Hare, and Simperl}]{VOUGIOUKLIS20181}
Pavlos Vougiouklis, Hady Elsahar, Lucie-Aimée Kaffee, Christophe Gravier,
  Frédérique Laforest, Jonathon Hare, and Elena Simperl. 2018.
\newblock \href {https://doi.org/https://doi.org/10.1016/j.websem.2018.07.002}
  {Neural wikipedian: Generating textual summaries from knowledge base
  triples}.
\newblock \emph{Journal of Web Semantics}, 52-53:1 -- 15.

\bibitem[{Wadden et~al.(2019)Wadden, Wennberg, Luan, and
  Hajishirzi}]{wadden-etal-2019-entity}
David Wadden, Ulme Wennberg, Yi~Luan, and Hannaneh Hajishirzi. 2019.
\newblock \href {https://doi.org/10.18653/v1/D19-1585} {Entity, relation, and
  event extraction with contextualized span representations}.
\newblock In \emph{Proceedings of the 2019 Conference on Empirical Methods in
  Natural Language Processing and the 9th International Joint Conference on
  Natural Language Processing (EMNLP-IJCNLP)}, pages 5784--5789, Hong Kong,
  China. Association for Computational Linguistics.

\bibitem[{Wang et~al.(2020)Wang, Wan, and Jin}]{doi:10.116200297}
Tianming Wang, Xiaojun Wan, and Hanqi Jin. 2020.
\newblock \href {https://doi.org/10.1162/tacl\_a\_00297} {Amr-to-text
  generation with graph transformer}.
\newblock \emph{Transactions of the Association for Computational Linguistics},
  8:19--33.

\bibitem[{Wiseman et~al.(2017)Wiseman, Shieber, and
  Rush}]{wiseman-etal-2017-challenges}
Sam Wiseman, Stuart Shieber, and Alexander Rush. 2017.
\newblock \href {https://doi.org/10.18653/v1/D17-1239} {Challenges in
  data-to-document generation}.
\newblock In \emph{Proceedings of the 2017 Conference on Empirical Methods in
  Natural Language Processing}, pages 2253--2263, Copenhagen, Denmark.
  Association for Computational Linguistics.

\bibitem[{Wolf et~al.(2019)Wolf, Debut, Sanh, Chaumond, Delangue, Moi, Cistac,
  Rault, Louf, Funtowicz, and Brew}]{wolf2019huggingfaces}
Thomas Wolf, Lysandre Debut, Victor Sanh, Julien Chaumond, Clement Delangue,
  Anthony Moi, Pierric Cistac, Tim Rault, Rémi Louf, Morgan Funtowicz, and
  Jamie Brew. 2019.
\newblock \href {http://arxiv.org/abs/1910.03771} {Huggingface's transformers:
  State-of-the-art natural language processing}.

\bibitem[{Yang et~al.(2019{\natexlab{a}})Yang, Dai, Yang, Carbonell,
  Salakhutdinov, and Le}]{NEURIPS2019_dc6a7e65}
Zhilin Yang, Zihang Dai, Yiming Yang, Jaime Carbonell, Russ~R Salakhutdinov,
  and Quoc~V Le. 2019{\natexlab{a}}.
\newblock \href
  {https://proceedings.neurips.cc/paper/2019/file/dc6a7e655d7e5840e66733e9ee67cc69-Paper.pdf}
  {Xlnet: Generalized autoregressive pretraining for language understanding}.
\newblock In \emph{Advances in Neural Information Processing Systems},
  volume~32, pages 5753--5763. Curran Associates, Inc.

\bibitem[{Yang et~al.(2019{\natexlab{b}})Yang, Dai, Yang, Carbonell,
  Salakhutdinov, and Le}]{NIPS2019_8812}
Zhilin Yang, Zihang Dai, Yiming Yang, Jaime Carbonell, Russ~R Salakhutdinov,
  and Quoc~V Le. 2019{\natexlab{b}}.
\newblock \href
  {http://papers.nips.cc/paper/8812-xlnet-generalized-autoregressive-pretraining-for-language-understanding.pdf}
  {Xlnet: Generalized autoregressive pretraining for language understanding}.
\newblock In H.~Wallach, H.~Larochelle, A.~Beygelzimer, F.~d\textquotesingle
  Alch\'{e}-Buc, E.~Fox, and R.~Garnett, editors, \emph{Advances in Neural
  Information Processing Systems 32}, pages 5753--5763. Curran Associates, Inc.

\bibitem[{Yao et~al.(2020)Yao, Wang, and Wan}]{yao-etal-2020-heterogeneous}
Shaowei Yao, Tianming Wang, and Xiaojun Wan. 2020.
\newblock \href {https://doi.org/10.18653/v1/2020.acl-main.640} {Heterogeneous
  graph transformer for graph-to-sequence learning}.
\newblock In \emph{Proceedings of the 58th Annual Meeting of the Association
  for Computational Linguistics}, pages 7145--7154, Online. Association for
  Computational Linguistics.

\bibitem[{Yu et~al.(2019)Yu, Zhang, Er, Li, Xue, Pang, Lin, Tan, Shi, Li,
  Jiang, Yasunaga, Shim, Chen, Fabbri, Li, Chen, Zhang, Dixit, Zhang, Xiong,
  Socher, Lasecki, and Radev}]{yu-etal-2019-cosql}
Tao Yu, Rui Zhang, Heyang Er, Suyi Li, Eric Xue, Bo~Pang, Xi~Victoria Lin,
  Yi~Chern Tan, Tianze Shi, Zihan Li, Youxuan Jiang, Michihiro Yasunaga,
  Sungrok Shim, Tao Chen, Alexander Fabbri, Zifan Li, Luyao Chen, Yuwen Zhang,
  Shreya Dixit, Vincent Zhang, Caiming Xiong, Richard Socher, Walter Lasecki,
  and Dragomir Radev. 2019.
\newblock \href {https://doi.org/10.18653/v1/D19-1204} {{C}o{SQL}: A
  conversational text-to-{SQL} challenge towards cross-domain natural language
  interfaces to databases}.
\newblock In \emph{Proceedings of the 2019 Conference on Empirical Methods in
  Natural Language Processing and the 9th International Joint Conference on
  Natural Language Processing (EMNLP-IJCNLP)}, pages 1962--1979, Hong Kong,
  China. Association for Computational Linguistics.

\bibitem[{Zhang et~al.(2020)Zhang, Kishore, Wu, Weinberger, and
  Artzi}]{bert-score}
Tianyi Zhang, Varsha Kishore, Felix Wu, Kilian~Q. Weinberger, and Yoav Artzi.
  2020.
\newblock \href {https://openreview.net/forum?id=SkeHuCVFDr} {Bertscore:
  Evaluating text generation with bert}.
\newblock In \emph{International Conference on Learning Representations}.

\bibitem[{Zhao et~al.(2020{\natexlab{a}})Zhao, Walker, and
  Chaturvedi}]{zhao-etal-2020-bridging}
Chao Zhao, Marilyn Walker, and Snigdha Chaturvedi. 2020{\natexlab{a}}.
\newblock \href {https://doi.org/10.18653/v1/2020.acl-main.224} {Bridging the
  structural gap between encoding and decoding for data-to-text generation}.
\newblock In \emph{Proceedings of the 58th Annual Meeting of the Association
  for Computational Linguistics}, pages 2481--2491, Online. Association for
  Computational Linguistics.

\bibitem[{Zhao et~al.(2019)Zhao, Peyrard, Liu, Gao, Meyer, and
  Eger}]{zhao-etal-2019-moverscore}
Wei Zhao, Maxime Peyrard, Fei Liu, Yang Gao, Christian~M. Meyer, and Steffen
  Eger. 2019.
\newblock \href {https://doi.org/10.18653/v1/D19-1053} {{M}over{S}core: Text
  generation evaluating with contextualized embeddings and earth mover
  distance}.
\newblock In \emph{Proceedings of the 2019 Conference on Empirical Methods in
  Natural Language Processing and the 9th International Joint Conference on
  Natural Language Processing (EMNLP-IJCNLP)}, pages 563--578, Hong Kong,
  China. Association for Computational Linguistics.

\bibitem[{Zhao et~al.(2020{\natexlab{b}})Zhao, Chen, Chen, Cao, Zhu, and
  Yu}]{zhao-etal-2020-line}
Yanbin Zhao, Lu~Chen, Zhi Chen, Ruisheng Cao, Su~Zhu, and Kai Yu.
  2020{\natexlab{b}}.
\newblock \href {https://doi.org/10.18653/v1/2020.acl-main.67} {Line graph
  enhanced {AMR}-to-text generation with mix-order graph attention networks}.
\newblock In \emph{Proceedings of the 58th Annual Meeting of the Association
  for Computational Linguistics}, pages 732--741, Online. Association for
  Computational Linguistics.

\bibitem[{Zhu et~al.(2019)Zhu, Li, Zhu, Qian, Zhang, and
  Zhou}]{zhu-etal-2019-modeling}
Jie Zhu, Junhui Li, Muhua Zhu, Longhua Qian, Min Zhang, and Guodong Zhou. 2019.
\newblock \href {https://doi.org/10.18653/v1/D19-1548} {Modeling graph
  structure in transformer for better {AMR}-to-text generation}.
\newblock In \emph{Proceedings of the 2019 Conference on Empirical Methods in
  Natural Language Processing and the 9th International Joint Conference on
  Natural Language Processing (EMNLP-IJCNLP)}, pages 5459--5468, Hong Kong,
  China. Association for Computational Linguistics.

\end{thebibliography}
\bibliographystyle{acl_natbib}

\clearpage

\appendix

\section*{Appendices}

In this supplementary material, we provide: (i) additional information about the data used in the experiments, and (ii) results that we could not fit into the main body of the paper.

\section{AMR Input Representation}
\label{sec:amrinput}
We test three variants for the representation of the input AMR graph. Following previous work~\cite{konsas_17, mager2020gpttoo}, we evaluate (i) only node representation, where the edge information is removed from the linearization; (ii) depth-first search (DFS) through the graph and the (iii) \textsc{penman} representation. An example for each representation is illustrated below: 
\begin{table}[h]
\begin{tabular}{c p{5.5cm}}
\vspace{3mm}
\small only nodes   & \small \texttt{value interrogative commodity true}\\
\small DFS     & \small \texttt{value :mode interrogative :ARG1 commodity :ARG1-of true} \\
\small \textsc{penman}     & \small \texttt{( value :mode interrogative :ARG1 ( commodity ) :ARG1-of ( true ) )}
\end{tabular}
\end{table}

In this experiment we employ T5\textsubscript{small}. Table~\ref{tab:amrinputs} shows the results on the AMR development set. The \textsc{penman} representation leads to best results. Therefore, this representation is used in the rest of the experiments.

\begin{table}[h]
\centering
{\renewcommand{\arraystretch}{0.8}

\begin{tabular}{lc}  
\toprule
 \textbf{Input} & \textbf{BLEU}  \\
 \midrule
 only nodes & 28.22 \\
 DFS & 34.94 \\
 \textsc{penman} & 38.27 \\
\bottomrule
\end{tabular}}
\caption{Results on the AMR dev set using T5\textsubscript{small} for different AMR linearizations.}
\label{tab:amrinputs}
\end{table}
\vspace{-4mm}

\section{Cross-domain Adaptation}
\label{sec:crossdomain}

For a given task, it is not always possible to collect closely related data -- as we saw, e.g., for WebNLG.
We therefore report \textsc{sta} in a cross-domain setting for the different KG-to-text benchmarks.
Table~\ref{tab:crossdomain_adddata} shows the results using BART\textsubscript{base} and T5\textsubscript{base}. While the texts in KGAIA and AGENDA share the domain of scientific abstracts, texts in WebNLG are more general. Also note that WebNLG graphs do not share any relations with the other KGs. For BART\textsubscript{base}, \textsc{sta} increases the performance in the cross-domain setting in most of the cases. For T5\textsubscript{base}, \textsc{sta} in KGAIA improves the performance on WebNLG.

In general, we find that exploring additional adaptive pretraining for graph-to-text generation can improve the performance even if the data do not come from the same domain.

\begin{table}[h]
\centering
{\renewcommand{\arraystretch}{0.8}
\begin{tabular}{lcc}  
\toprule
\textbf{STA on} & \multicolumn{2}{c}{\textbf{Fine-tuned \&{} Evaluated on}} \\
\cmidrule(lr){2-3}
 & WebNLG-\textit{Seen} & AGENDA \\
\midrule
\multicolumn{3}{c}{BART\textsubscript{base}} \\
\midrule
None & 58.71 & 22.01 \\
KGAIA & 63.20 & 23.48 \\
WebNLG & - & 21.98 \\
AGENDA & 61.25 & - \\

\midrule
\multicolumn{3}{c}{T5\textsubscript{base}} \\
\midrule
None & 62.93 & 20.73 \\
KGAIA & 63.19 & 22.44 \\
WebNLG & - & 20.27 \\
AGENDA & 62.75 & - \\

\bottomrule
\end{tabular}}
\caption{Effect (measured with BLEU score) of cross-domain \textsc{sta}.}
\label{tab:crossdomain_adddata}
\end{table}
\vspace{-5mm}

\section{Input Graph Size}
\label{section:inputgraphsize}
\begin{figure}[h]
    \centering
    \includegraphics[width=.3\textwidth]{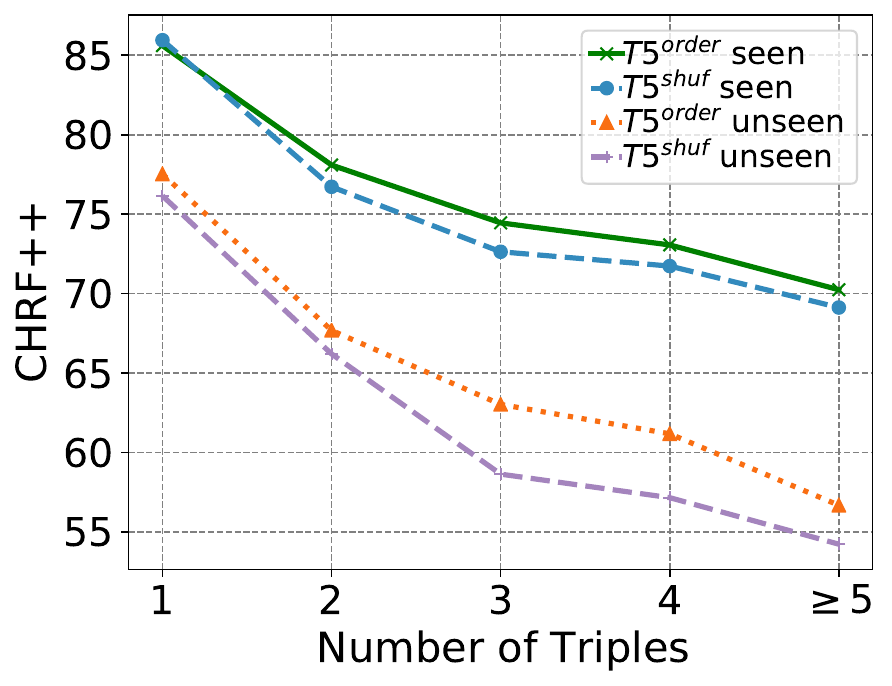}
    \caption{chrF++ scores with respect to the number of triples for WebNLG \textit{seen} and \textit{unseen} test sets.}
    \label{fig:graphs-triples}
\end{figure}

Figure~\ref{fig:graphs-triples} visualizes T5\textsubscript{small}'s performance with respect to the number of input graph triples in WebNLG dataset.
We observe that \shufmodel{T5}{small}{order} and \shufmodel{T5}{small}{shuf} perform similarly for inputs with only one triple but that the gap between the models increases with larger graphs. While it is obviously more difficult to reconstruct a larger graph than a smaller one, this also suggests that the graph structure is more taken into account for graphs with more than 2 triples.
For the \textit{unseen} setting, the performance gap for these graphs is even larger, suggesting that the PLM can make more use of the graph structure when it has to.

\vspace{-6mm}

\begin{table*}[t]
\centering
{\renewcommand{\arraystretch}{0.8}
\begin{tabular}{lp{0.2cm}rrrp{0.2cm}rrrp{0.2cm}rrrp{0.2cm}}  
\toprule
& & \multicolumn{3}{c}{\textbf{AMR}} & & \multicolumn{3}{c}{\textbf{WebNLG}} & & \multicolumn{3}{c}{\textbf{AGENDA}} \\
\midrule
min, avg and max number of nodes & & 2 & 28.6 & 335 &  & 2 & 6.8 & 15 & & 2 & 10.5 & 80 \\
min, avg and max node degrees & & 1 & 2.2 & 21 & & 1 & 1.7 & 7 & & 1 & 1.67 & 15 \\
min, avg and max number of edges & & 1 & 32.3 & 554 & & 1 & 5.9 & 14 & & 1 & 8.8 & 124 \\
min, avg and max graph diameter & & 1 & 12.2 & 40 & & 1 & 4.1 & 10 & & 1 & 3.1 & 20\\
min, avg and max shortest path length & & 0 & 7.49 & 40 & & 0 & 2.4 & 10 & & 0 & 2.3 & 20\\
\bottomrule
\end{tabular}}
\caption{Graph statistics of AMR, WebNLG and AGENDA datasets. The values are calculated using the training data. Note that AMR graphs contain a more complex structure than WebNLG and AGENDA graphs.}
\label{tab:graphstats}
\end{table*}

\section{Graph Statistics}
\label{sec:graphstats}

In Table~\ref{tab:graphstats}, we present the graph properties of the three datasets. All statistics are calculated using the Levi transformation \cite{beck-etal-2018-acl2018} of the undirected version of the graphs, where edges are also considered nodes in the graph. WebNLG and AGENDA datasets contain disconnected graphs, and we use the largest subgraph to calculate the diameter. Note that AMR graphs have a much more complex structure: (i) they have more nodes and edges than WebNLG and AGENDA graphs; (ii) the average graph diameter and the average shortest path between nodes in AMRs are at least three times larger than in WebNLG and AGENDA graphs; (iii) nodes in AMRs have larger degrees than nodes in WebNLG and AGENDA graphs.

\vspace{15mm}

\begin{table}[h]
{\renewcommand{\arraystretch}{0.8}
\begin{tabular}{@{\hspace*{1mm}}l@{\hspace*{2mm}}r@{\hspace*{3mm}}r@{\hspace*{2mm}}r@{\hspace*{2mm}}r@{\hspace*{1mm}}}  
\toprule
 & \textbf{AMR17} & \textbf{WebNLG} & \textbf{AGENDA}  \\
\midrule
\#Train & 36,521 & 18,102 & 38,720  \\
\#Dev & 1,368 & 872 & 1,000  \\
\#Test & 1,371 & 1,862 & 1,000 \\
\midrule
\#Relations & 155 & 373 & 7 \\
Avg \#Tokens & 16.1 & 31.5 & 157.9 \\

\bottomrule
\end{tabular}}
    \caption{Statistics for the graph-to-text benchmarks.}
    \label{tab:datastatistics}
\end{table}
\begin{table}[h]
{\renewcommand{\arraystretch}{0.8}
\begin{tabular}{@{\hspace*{4mm}}l@{\hspace*{6mm}}c@{\hspace*{6mm}}c@{\hspace*{6mm}}c@{\hspace*{3mm}}c@{\hspace*{14mm}}}  
\toprule
 & \textbf{Title} & \textbf{Abstract} & \textbf{KG}  \\
 \midrule
 Vocab & 48K & 173K & 113K \\
 Tokens & 2.1M & 31.7M & 9.6M \\
 Entities & - & - & 3.7M \\
 Avg Length & 11.1 & 167.1 & - \\
 Avg \#Nodes & - & - & 19.9 \\
 Avg \#Edges & - & - & 9.4 \\

\bottomrule
\end{tabular}}
\caption{Statistics for the KGAIA dataset.}
\label{tab:augstatistics}
\end{table}

\begin{table}[h]
{\renewcommand{\arraystretch}{0.6}
\begin{tabular}{@{\hspace*{1mm}}l@{\hspace*{1mm}}ccc@{\hspace*{1mm}}}  
\toprule
\textbf{Model} & \textbf{chrF++} & \textbf{BS (F1)} & \textbf{MS}  \\
\midrule
\citet{schmitt2020modeling} & 44.53 & -& -\\
\citet{ribeiro-etal-2020-modeling} & 46.37 & -& - \\
\midrule
BART\textsubscript{base} & 48.02 & 89.36 & 34.33 \\
BART\textsubscript{large} & \textbf{50.44} & 88.74 & 32.24 \\
T5\textsubscript{small}  & 44.91 & 88.56 & 30.25 \\
T5\textsubscript{base} & 48.14 &  88.81 & 31.33 \\
T5\textsubscript{large} &  48.14 & \textbf{89.60} & \textbf{35.23} \\
\midrule
\multicolumn{2}{l}{\small{\textit{with task-adaptive pretraining}}}  \\[.2em]
BART\textsubscript{large} + \textsc{lma} & 51.33 & 89.12 & 33.42 \\
T5\textsubscript{large} + \textsc{lma} & 49.37 & 89.75 & 36.13 \\[.7em]
BART\textsubscript{large} + \textsc{sta} & \textbf{\textit{51.63}} & 89.27 & 34.28 \\
T5\textsubscript{large} + \textsc{sta} & 50.27 & \textbf{\textit{89.93}} & \textbf{\textit{36.86}} \\
\bottomrule
\end{tabular}}
\caption{Results of the chrF++, BertScore (BS) and MoverScore (MS) scores for AGENDA test set. \textbf{Bold} (\textbf{\textit{Italic}}) indicates best scores without (with) task-adaptive pretraining.}
\label{tab:results-agenda-appendix}
\end{table}

\begin{table}[t]
{\renewcommand{\arraystretch}{0.6}
\begin{tabular}{@{\hspace*{1mm}}l@{\hspace*{1mm}}c@{\hspace*{2mm}}c@{\hspace*{2mm}}c@{\hspace*{1mm}}}  
\toprule
\textbf{Model} & \textbf{chrF++} & \textbf{BS (F1)} & \textbf{MS}  \\
\midrule
\citet{dcgcnforgraph2seq19guo} & 57.30 & - & - \\
\citet{zhu-etal-2019-modeling} & 64.05 & - & - \\
\citet{cai-lam-2020-graph} & 59.40 & - & - \\
\citet{doi:10.116200297} &  65.80 & - & - \\
\citet{yao-etal-2020-heterogeneous}  &  65.60 & - & - \\
\midrule
\small{\textit{based on PLMs}}  &  \\[.2em]
\citet{mager2020gpttoo}  & 63.89  & -& - \\
\midrule
BART\textsubscript{base}  &  66.65 & 95.22 & 60.78  \\
BART\textsubscript{large}  & 71.06 & 96.08 & 65.74  \\
T5\textsubscript{small} &  68.78 & 95.62 & 63.70  \\
T5\textsubscript{base} & 70.81 & 95.99 & 65.63  \\
T5\textsubscript{large} & \textbf{72.57} & \textbf{96.27} & \textbf{67.37} \\
\midrule
\multicolumn{3}{l}{\small{\textit{with task-adaptive pretraining}}}  \\[.2em]
BART\textsubscript{large} + \textsc{lma} & 71.14 & 95.94 & 64.75  \\
T5\textsubscript{large} + \textsc{lma} & 72.83 & 96.32 & 67.44  \\[.7em]
BART\textsubscript{large} + \textsc{sta} \small{\textsc{(200K)}} & 72.26 & 96.21 &66.75 \\
BART\textsubscript{large} + \textsc{sta} \small{\textsc{(2M)}} & 73.58 & 96.43 & 68.14 \\
T5\textsubscript{large} + \textsc{sta} \small{\textsc{(200K)}} & 74.09 & 96.51 & 68.86 \\
T5\textsubscript{large} + \textsc{sta} \small{\textsc{(2M)}} & \textbf{\textit{74.79}} & \textbf{\textit{96.59}} & \textbf{\textit{69.53}} \\
\bottomrule
\end{tabular}}
\caption{Results of the chrF++, BertScore (BS) and MoverScore (MS) scores for the LDC2017T10 test set. \textbf{Bold} (\textbf{\textit{Italic}}) indicates the best score without (with) task-adaptive pretraining.}
\label{tab:results-amr-appendix}
\end{table}



\begin{table*}[t]
    \centering
    \footnotesize
    \begin{tabular}{@{\hspace{.1em}}c@{\hspace{.1em}}c@{\hspace{.8em}}p{5cm}@{\hspace{.8em}}p{4.5cm}@{\hspace{.8em}}p{4.5cm}@{\hspace{.8em}}}
    \toprule
         &\textbf{T/F}&\multicolumn{1}{c}{\textbf{Input Facts}} & \multicolumn{1}{c}{\textbf{\shufmodel{T5}{small}{order}}} & \multicolumn{1}{c}{\textbf{\shufmodel{T5}{small}{shuf}}}  \\
         \midrule
        (1) &S& \sep{} capital \sep{} leader Name \sep{} London \sep{} Pound sterling \sep{} United Kingdom \sep{} leader Name \sep{} United Kingdom \sep{} Elizabeth II \sep{} United Kingdom \sep{} Boris Johnson \sep{} London \sep{} currency & The capital city is London, the currency is the Pound sterling and the leader is Elizabeth II. Boris Johnson is also a leader in the UK. & The capital of the United Kingdom is London, the currency is the Pound sterling and the country is lead by Elizabeth II and Boris Johnson.\\
         \midrule
                 (2) &T& \sep{} Germany \sep{} capital \sep{} Berlin & Berlin is the capital of \mbox{Germany.}  & Berlin is the capital of \mbox{Germany.}  \\
                 \midrule
         (3) &F& \sep{} Berlin \sep{} capital \sep{} Germany & Berlin's capital is \mbox{Germany.}  & Berlin is the capital of \mbox{Germany.}  \\
         
         \midrule
                 (4) &F& \sep{} Leinster \sep{} is Part Of \sep{} Dublin & Leinster is part of Dublin. & Leinster is part of Dublin. \\
         \midrule
         (5) &F& \sep{} Rome \sep{} capital \sep{} Italy & Rome’s capital is Italy. & Rome is the capital of Italy. \\
         \midrule
         (6) &T& \sep{} Italy \sep{} capital \sep{} Rome & Italy’s capital is Rome. & Rome is the capital of Italy.\\
         \midrule
          (7) &T& \sep{} Texas \sep{} capital \sep{} Austin \sep{} Andrews County Airport \sep{} location \sep{} Texas & Austin is the capital of Texas where Andrews County Airport is located.  &  Austin is the capital of Texas where Andrews County Airport is located. \\
          \midrule
         (8) &F& \sep{} Austin \sep{} capital \sep{} Texas \sep{} Andrews County Airport \sep{} location \sep{} Texas  & The capital of Austin is Texas and Andrews County Airport is located in Texas. & Andrews County Airport is located in Texas where Austin is the capital.
  \\
         \bottomrule
    \end{tabular}
    \caption{Example generations from shuffled (S), true (T), and corrupted (F) triple facts by T5\textsubscript{small}, fine-tuned on correctly ordered triples (\emph{order}) and randomly shuffled input (\emph{shuf}).}
    \label{tab:qualitative_appendix}
\end{table*}
\begin{table*}[h]
\begin{center}
\begin{tabular}{llp{11cm}}
\toprule
\textbf{D}     & \textbf{Model}       & \textbf{Examples}    \\
\midrule
\vertmulticell{10}{AMR}  & Reference  & I had to deal with verbal abuse from my dad for a long 8 years before I came to uni and honestly, the only reason why I'm here is because it was the only way out. \\
 & T5 & I had to deal with 8 years of verbal abuse from my dad before coming to university and honestly the only reason I'm here is because it's the only way out. \\
 & BART & I had to deal with my dad's verbal abuse for 8 years long before coming to uni and honestly the only reason I'm here is because it's the only way out. \\
 & \citet{mager2020gpttoo} & i've had to deal with verbal abuse from my dad for 8 years (before i came to uni i was honestly the only reason i was here) and it's only because of the way it is. \\
 
 \midrule
\vertmulticell{10}{WebNLG}  & Reference & Aaron Turner is an electric guitar player who has played with the black metal band Twilight and with Old Man Gloom. Death metal is a musical fusion of black metal.  \\
    & T5 & Aaron Turner plays the electric guitar and is associated with the band Twilight. He is also a member of the Old Man Gloom band. Black metal and death metal are both genres of music. \\
       & BART & The black metal genre is a variation of the death metal genre. It is also where the band, Twilight, and the alternative rock band, Old Man Gloom, are from. One of the members of the band is Aaron Turner, who plays the electric guitar. \\
   & \citet{harkous2020text} & Aaron Turner, a.k.a. Black Metal, is a member of the Twilight (band) and Old Man Gloom bands. He also plays electric guitar and has a strong connection with the Death Metal genre. \\
\midrule
\end{tabular} 
\caption{Examples of text generated by the different models. D refers to the dataset.}\label{tab:human_examples}
\end{center}
\end{table*} 

\begin{table*}[h]
\begin{center}
\begin{tabular}{lp{12cm}}
\toprule
 \textbf{Model}       & \textbf{Examples}    \\
\midrule
    Reference &  in this paper, a new array signal processing technique by using particle swarm optimization is proposed to identify multipath channel parameters. the proposed array signal processing technique provides estimates to the channel parameters by finding a global minimum of an optimization problem. since the optimization problem is formulated in the cross-ambiguity function (caf) domain of the transmitted signal and the received array outputs, the proposed array signal processing technique is called as particle swarm optimization. the performance of the particle swarm optimization is compared with the space alternating generalized expectation maximization technique and with another recently proposed pso based technique for various snr values. simulation results indicate the superior performance of the pso based technique over mentioned techniques for all snr values. \\
    \midrule
 T5  & this paper, we propose an array signal processing technique based on particle swarm optimization to estimate the multipath channel parameters from the received and the transmitted signal in a cross-ambiguity domain. the proposed array signal processing technique is formulated as an optimization problem and particle swarm optimization is compared with the generalized expectation maximization technique. simulation results show that the proposed particle swarm optimization outperforms the pso based technique for a wide range of snr values. in addition, the proposed particle swarm optimization is computationally more efficient than the gem based technique for small snr values. \\
 \midrule
 BART  & in this paper, a new array signal processing technique based on particle swarm optimization is proposed. the proposed array signal processing technique is used to estimate the multipath channel parameters from the transmitted signal. the proposed array signal processing technique is formulated as an optimization problem in the cross-ambiguity domain. the particle swarm optimization is compared with the generalized expectation maximization technique. simulation results show that the proposed particle swarm optimization outperforms the pso based technique for all snr values. furthermore, the proposed particle swarm optimization is able to estimate the channel parameters more accurately than the generalized expectation maximization technique. \\
 \midrule
  \citet{ribeiro-etal-2020-modeling}  & in this paper, a novel array signal processing technique based on particle swarm optimization is proposed to estimate the multipath channel parameters from the transmitted signal. the proposed array signal processing technique uses particle swarm optimization to estimate the multipath channel parameters. the proposed array signal processing technique is formulated as an optimization problem. simulation results show that the proposed array signal processing technique outperforms the conventional generalized expectation maximization technique and the pso based technique is robust to the snr values. \\

\midrule
\end{tabular} 
\caption{Examples of text generated by the different models trained on the AGENDA dataset.}\label{tab:human_examples_agenda}
\end{center}
\end{table*}

\end{document}